\documentclass[a4paper]{siamart1116}

\usepackage{times}

\newtheorem{defn}{Definition}
\newtheorem{assump}{Assumption}
\newtheorem{prop}{Proposition}

\usepackage{siam-water-private}

\usepackage{amsopn} 

\DeclareMathOperator{\rank}{rank}
\DeclareMathOperator{\range}{columnspace}
\DeclareMathOperator{\rowspace}{rowspace}

\DeclareMathOperator{\cokernel}{cokernel}

\DeclareMathOperator*{\argmin}{arg\,min}

\usepackage{amsfonts}
\usepackage{graphicx}
\usepackage{algorithmic}
\ifpdf
  \DeclareGraphicsExtensions{.pdf,.png,.jpg}
\else
  \DeclareGraphicsExtensions{.eps}
\fi

\newcommand{\TheTitle}{Overcoming model simplifications when quantifying predictive uncertainty} 
\newcommand{\TheAuthors}{G. M. Mathews and J. Vial}

\headers{Overcoming model simplifications}{\TheAuthors}

\title{{\TheTitle}%
}

\author{
  George M. Mathews\thanks{Data61, CSIRO, Australia}
  \and
  John Vial\footnotemark[1]
}

\pdfoutput=1

\ifpdf
\hypersetup{
  pdftitle={\TheTitle},
  pdfauthor={\TheAuthors}
}
\fi

\begin{document}

\maketitle

\begin{abstract}
It is generally accepted that all models are wrong -- the difficulty is 
determining which are useful. Here, a useful model is considered as one that 
is capable of combining data and expert knowledge, through an inversion or 
calibration process, to adequately characterize the 
uncertainty in predictions of interest. 
This paper derives conditions that specify which simplified models are useful and how they should be calibrated.  
To start, the notion of an optimal simplification is defined. This relates the model 
simplifications to the nature of the data and predictions, and determines when 
a standard probabilistic calibration scheme is capable of accurately 
characterizing uncertainty.
Furthermore, two additional conditions are defined for suboptimal models that 
determine when the simplifications can be safely ignored. 
The first allows a suboptimally simplified model to be used in a way that 
replicates the performance of an optimal model. This is achieved through the 
judicial selection of a prior term for the calibration process that explicitly 
includes the nature of the data, predictions and modelling simplifications. 
The second considers the dependency structure between the predictions and the 
available data to gain insights into when the simplifications can be overcome 
by using the right calibration data. Furthermore, the derived 
conditions are related to the commonly used calibration schemes based on 
Tikhonov and subspace regularization. To allow concrete insights to be obtained, 
the analysis is performed under a linear 
expansion of the model equations and where the predictive uncertainty is 
characterized via second order moments only. 

\end{abstract}


\section{Introduction}

This paper considers the problem of assessing uncertainty in a prediction made for a particular system based on the combination of specific measurement data and expert domain knowledge. The focus of this work is environmental systems, such as river basins and groundwater systems, however, the analysis is likely to be applicable to a much wider set of domains. Correctly addressing such environmental prediction problems is fundamental to ensuring these important system are managed appropriately and sustainabl. Probabilistic Bayesian methods provide a theoretically consistent set of rules to combine such site specific data and prior knowledge through the use of a system model. However, these methods often fail when naively applied to a model that does not capture the full complexity of the system.

A possible solution  is the incorporation of more detail and structural variations in the system model such that more sources of uncertainty are included.  However, it is important to admit that this modelling \textit{ad.~infinitum} is not a solution as unlimited resources are never available for detailed model construction and execution, and simplifications must be made, at least at some level. 

However, deciding on what should be included in a model of a system and what may be ignored is not straightforward. For instance in groundwater hydrology there is no consensus on what is an appropriate level of parameterization detail, e.g within the hydraulic conductivity field \cite{hunt_are_2007, voss_editors_2011, voss_editors_2011-1, white_quantifying_2014}. In addition to this are the related decisions of what processes should be explicitly represented and what can be ignored. Or alternatively, when is it reasonable to lump many different processes together under a semi-physical or non-physical black box model that directly represents input-output relationships \cite{ferdowsian_explaining_2001, von_asmuth_modeling_2008}.

What is necessary is an understanding of the effects of simplifying assumptions made during model development, and appropriate ways of dealing with them when calibrating the model and generating predictions. 

A general, qualitative, notion of model adequacy was explored by \citet{gupta_towards_2012} in terms of the issues faced within the surface water, groundwater, unsaturated zone, and terrestrial hydrometeorology modelling communities. The work considered intermediate stages within the modelling process from the initial perceptual understanding through to the construction of a computational model and introduced a pluralistic definition of model adequacy. On one extreme is the engineering viewpoint that defines a structurally adequate model as one that can reproduce the input-output relationship of the system, with well characterized uncertainties (error models). On the other extreme is the physical science viewpoint that requires an adequate model to be consistent with the underlying physical system \cite{gupta_towards_2012}.

This paper will consider systems that may be exposed to future disturbances and limited data is available. Within these problems a well characterized regression model (the engineering viewpoint) cannot be constructed and some physical insight of the system is required \cite{gupta_towards_2012}. Furthermore, focus is given to environmental management problems, where the role of a model is to help inform a subsequent decision problem related to risk management e.g. engineering design \cite{freeze_hydrogeological_1990}, groundwater management \cite{doherty_groundwater_2013}, or climate change \cite{rougier_uncertainty_2014}. This requires the specification of a probability distribution over the predictions of interest such that decision theoretic methods can be applied to quantify the risk and determine the optimal management strategy \cite{berger_statistical_1985, smith_bayesian_2010}. 

\subsection{Related Work}

The issue of mismatch between reality and a numerical representation of a system has received considerable attention from may different perspectives. 
The presence of simplifications may be identified in the data as additional misfit that is not consistent with measurements errors alone. These additional errors are referred to as: model error \cite{mclaughlin_reassessment_1996, white_quantifying_2014, ljung_stochastic_2014}, model structural error \cite{beven_concept_2005}, structural noise \cite{doherty_short_2010}, model inadequacy \cite{kennedy_bayesian_2001, gupta_towards_2012}, model discrepancy \cite{goldstein_probabilistic_2004, strong_when_2014}, modelization uncertainties \cite{tarantola_inverse_2005}, and others.

Within the system identification and control community \cite{ljung_system_1999, ljung_perspectives_2010} the concept is often referred to as system under-modelling \cite{ninness_estimation_1995, reinelt_comparing_2002} and two main probabilistic approaches have been developed: stochastic embedding \cite{goodwin_stochastic_1989, goodwin_quantifying_1992, ljung_stochastic_2014} and model error modelling \cite{ljung_model_1999}. These allow subjective information about the discrepancy between the transfer functions that describe the dynamics of the real system and that of a numerical model to be explicitly defined and combined with a time series dataset collected from the system. This in turn allows prediction uncertainty to include uncertainty due to modelling errors and measurement errors contained within the data. These approaches generally allow a very flexible black box representation of the system dynamics to be used.

In the statistical modelling community,  computational simulators have been considered as an explicit approximation to the real physical system they are attempting to represent. The mismatch is represented as an additional error term and modelled probabilistically, generally with explicit space time correlation \cite{kennedy_bayesian_2001, craig_bayesian_2001, goldstein_probabilistic_2004, goldstein_bayes_2006, rougier_probabilistic_2007, rougier_uncertainty_2014}. Due to the computational complexity of the methods for high dimensional models, approximate Bayesian methods have also been developed \cite{goldstein_bayes_2006}. Such simulators may also be decomposed and internal error terms included to represent the structural errors within smaller submodels \cite{strong_when_2014}.

Other work within hydrology has examined model simplifications and how these effect the calibration and prediction processes. In particular it has been shown that model parameters that are designed to represent particular physical properties may be forced to undertake surrogate roles to compensate for the simplifications during the model calibration process. This surrogacy has the potential to introduce additional biases into the predictions that are unaccounted for by typical probabilistic methods. This has been analyzed by explicitly modelling the simplification process in \cite{mclaughlin_distributed_1988, doherty_short_2010, doherty_use_2011, watson_parameter_2013, white_quantifying_2014}. 

Furthermore, several strategies have been considered to overcome the approximations inherent in a simplified model.
This includes the generalized likelihood uncertainty estimation (GLUE) method \cite{beven_equifinality_2001} that modifies the data likelihood function such that less information is extracted from the data. For a dynamical system, the assumption of a deterministic dynamic transition function can be relaxed and a stochastic process or transition function used instead \cite{vrugt_improved_2005, clark_unraveling_2006} (such models are often referred to as data assimilation methods). Similarly the assumption that the model parameters are time invariant may be related such that they can change over time to better match the observed system behavior \cite{reichert_analyzing_2009}. 

It is noted that many of these methods rely on a noticeable discrepancy between the measured data and model predicted values. However, the adverse effects of model simplifications may not necessarily cause any additional misfit between the data and model output, and thus may go unnoticed during model calibration and prediction \cite{white_quantifying_2014}. Thus it is critical to have a theoretical understanding of model adequacy that goes beyond data fit.

\subsection{Approach and Contributions}

This paper considers a subjective Bayesian framework and formally defines what is an adequate or appropriate model by explicitly considering: (i) the nature of the predictions, (ii) the available data and prior knowledge, (iii) the model simplification strategy, and (iv) the calibration scheme. It is shown that these are intrinsically linked and performing model calibration should take into account the nature of the data, predictions and any inherent limitations of the simplified computational model. It is noted that this departs from classical guidance on calibration and inversion that focuses on data misfit only \cite{mclaughlin_reassessment_1996, tarantola_inverse_2005, aster_parameter_2012}. 

A dual model approach is used, where a ``reference'' \cite{mclaughlin_distributed_1988} or ``reality'' \cite{white_quantifying_2014} model is used to describe how the system is believed to function. Such a high fidelity reference model has been used in the past to characterize the performance of a simplified model. Here, the approach is extended to explicitly determine when and how a simplified model can be used to generate an accurate, or at least conservative, estimate of uncertainty in a prediction of interest.
To allow concrete insights to be produced, only linear(ized) problems are considered and the beliefs are restricted to be Gaussian, where the second order moment is sufficient to capture uncertainty. 

Within this framework, model simplifications are represented as a subspace projection that restricts the flexibility of the simplified model. This prevents the model from being able to fully represent the complexity of how the real system is believed to function. It is shown that if the model simplifications are ignored and standard probabilistic calibration methods are used, the model may generate overconfident and non-conservative predictions. Generally, this is avoided only when the simplification strategy is \emph{optimal}. The key contribution of this paper is the characterization of two new calibration and prediction schemes for suboptimally simplified models that avoid this under estimation of uncertainty. The first scheme allows a simplified model to be used in a way that replicates the performance of an optimal scheme through the appropriate modification of the prior, or regularization, term. In the second scheme,  focus is given on linking the predictions with the available data such that insight is gained into when the simplifications can be overcome by gathering the appropriate set of data. 

Specifically, this paper fully defines the general problem of assessing uncertainty with a simplified model in \Cref{sec:problem-definition}. \Cref{sec:optimal} summarizes the optimal benchmark solution based on the high fidelity reference model. The general representation of model simplifications based on linear projection is defined in \Cref{sec:model-simplifications}. A typical probabilistic calibration scheme is considered in \Cref{sec:naive-use} that ignores the effects of model simplifications within the calibration and prediction process. This section defines the concept of an optimal simplification that determines when such a scheme is adequate for assessing uncertainty. \Cref{sec:overcomming-simplifications} considers suboptimally simplified models and introduces two new calibration and prediction schemes, and defines the conditions when they are adequate for assessing predictive uncertainty. Discussions and directions for future research are covered in \Cref{sec:conclusion}.  A worked example involving a simplified groundwater prediction problem is given in \Cref{sec:example}.

\section{Problem Definition}
\label{sec:problem-definition}

Consider a system where a modeller is tasked with generating a probability distribution over a prediction of a specific feature of the system to inform a future decision process. For instance in a groundwater system, the prediction of interest may be the reduction in water levels that will be experienced by an aquifer due to an increase in future extractions. The prediction of interest will be denoted by the vector $\p\in\Re^{D_\p}$.

To aid the modeller, a limited number of measurements have been made on the system and will be represented by $\d\in\Re^{D_\d}$. Furthermore, it is considered that expert prior knowledge exists on how the system functions, for instance the physical processes involved, and allows the available data and prediction of interest to be related to the properties of the underlying system. 

This information shall be considered to define a high fidelity or reference model for the system with parameters denoted by the vector $\x\in\Re^{D_\x}$. Furthermore, the relationship between the data $\d$, predictions $\p$, and the underlying parameters $\x$ that the modeller believes exists shall be represented by the pair of equations
\begin{align}
 \d &= \GG(\x) + \text{errors}, \label{eq:nonlinear-data-equation}\\  
 \p &= \YY(\x) + \text{errors}. \label{eq:nonlinear-predict-equation}
\end{align}
It is considered that these functions include the operation of known physical laws, such as conservation of mass, energy and momentum, while the vector $\x$ includes a detailed representation of the forcing terms, initial conditions, material properties, etc. These requirements result in a very high dimensional vector $\x$ and complex functions $\GG$ and $\YY$. %

Furthermore, it is considered that epistemic uncertainty exists as to the value of $\x$ for the system under investigation. In addition, the errors on the right of  \cref{eq:nonlinear-data-equation} and \cref{eq:nonlinear-predict-equation} allow for the inclusion of additional uncertainty, for instance to capture measurement errors introduced by the data acquisition method, or uncertainty that may exist in how a physical process actually functions \cite{tarantola_inverse_2005}. 

This allows the modeller's complete belief about how the system behaves to be captured by three probability distribution functions%
\begin{equation}
p(\x), 
\quad 
p(\d|\x), 
\quad  
p(\p|\x).
\label{eq:complex-pdfs}
\end{equation}
These in turn can be used to generate a posterior distribution over the predictions of interest, denoted as $p(\p | \d)$, using standard rules of probability theory. %

It is noted that a major issue with pursuing this type of subjective Bayesian approach is the  requirement that these probability distributions must represent every detail that is believed to be present in the physical system. This is not possible and simplification must be made.

\subsection{Simplifications}

\begin{figure}[tb]%
\small\centering
\includegraphics[width=0.8\columnwidth]{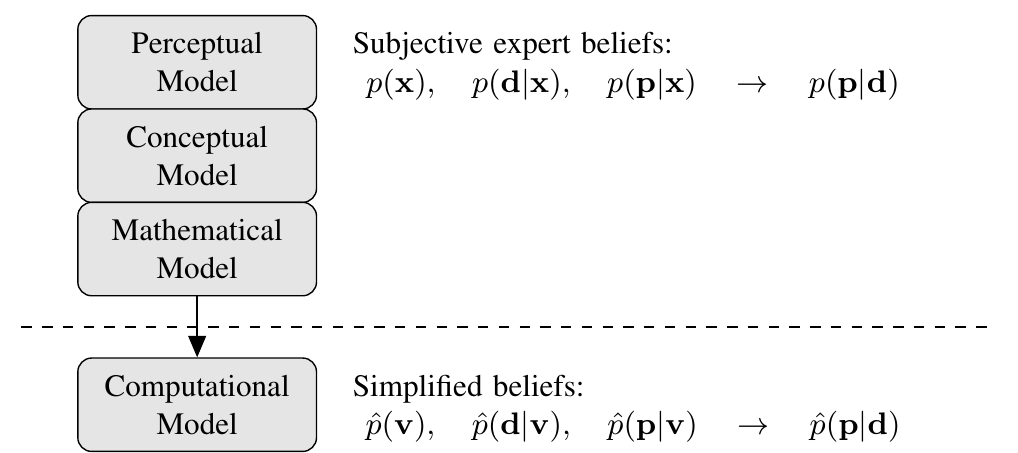}
\caption{The process of modelling is to transform the modeller's beliefs into a set of approximate probabilities that can be processed numerically to generate a conservative posterior probability distribution over the predictions of interest.}
\label{fig:modelling}
\end{figure}

It is considered that the process of modelling, for instance as outlined in \cite{gupta_towards_2012}, transforms the detailed knowledge of the modeller and produces the desired simplified computational representation of the problem that can be solved within the typical computational constraints. 
Specifically, the output of the modelling process is (or should be) an approximate representation of the modeller's beliefs that can be processed numerically to generate a prediction posterior distribution that is somehow similar to the optimal, but computationally intractable, distribution $p(\p | \d)$ generated by the high fidelity model. The modelling process is depicted in \cref{fig:modelling} where the outputs are denoted by the set of approximate beliefs 
\begin{equation}
\ap(\v), 
\quad 
\ap(\d|\v), 
\quad  
\ap(\p|\v).
\label{eq:simple-pdfs}
\end{equation}
These probability distributions are defined over a simplified description of the system, denoted by the parameter vector $\v\in\Re^{D_{\v}}$. 
Furthermore, it is considered that the approximate posterior $\ap(\p|\d)$ should not just be similar to posterior $p(\p | \d)$, it should be in some sense \emph{conservative} such that the uncertainty is not underestimated \cite{doherty_groundwater_2013}.

Thus, the key question that is addressed in this paper is now specified as: 

\vspace{2mm}
\emph{
\noindent How can the subjective beliefs of a modeller, defined for the high fidelity reference model, be transformed to generate approximate probabilities that define a simplified computational model such that the computed prediction posterior $\ap(\p|\d)$ is a conservative approximation of the optimal reference posterior $p(\p|\d)$? 
}
\vspace{2mm}

This question explicitly links the typically separate problems of: model construction, calibration and prediction. 

Furthermore, answers will be provided under the restriction that the uncertainties are described by multivariate Gaussian distributions and the dependencies are linear. This will allow more specific and concrete insights to be gained into the effects of simplifications and how they may be overcome. Extensions to more general non-Gaussian and nonlinear systems is left for future work.

\section{Optimal Inference and Prediction}
\label{sec:optimal}

The optimal Bayesian prediction scheme, which does not consider any simplifications, is now briefly reviewed \cite{berger_statistical_1985, tarantola_inverse_2005, aster_parameter_2012}. This is based on the high fidelity reference model.%


Let the predictions of interest, the available data, and the parameters of the high fidelity model be defined by the random vectors: $\p \in \Re^{D_\p}$, $\d \in \Re^{D_\d}$, and $\x \in \Re^{D_\x}$, where $D_\p$, $D_\d$ and $D_\x$ are their respective dimensions. It is considered that the dimension of $\x$ is excessively large, while the data is limited, and the number of predictions is small (perhaps only one), such that $D_\x \gg D_\d,D_\p$.

Prior knowledge of the parameter vector is represented by the Gaussian probability density function $p(\x)$, with mean of zero %
and known second order moment $\cov{\x}$
such that
\begin{equation}
p(\x) \define N(\x;0,\cov{\x}).
\label{eq:px}
\end{equation}
It is noted that the requirement of a zero mean distribution  simplifies the analysis and can be met in general with a transformation into increments.

Now, the believed relationship between the system properties $\x$, and the data vector $\d$ is considered to be linear with coefficient matrix $\G$. This can be considered as a linearized approximation of \cref{eq:nonlinear-data-equation}, however the approximation introduced by the linearization is considered out of scope in this paper. In addition, the uncertainty that is believed to exist in this relationship is represented by the error term $\nd$ such that $\d = \G \x + \nd$. Furthermore, the uncertainty in the value of $\nd$ is represented by a zero mean Gaussian density with known covariance matrix of $\cov{\nd}$. In the simplest case, the matrix $\cov{\nd}$ represents errors within the measurement process. This allows the modeller's conditional belief of the measured data given the underlying parameters to be defined by the Gaussian density
\begin{equation}
p(\d | \x) \define N(\d; \G\x, \cov{\nd}).
\label{eq:pd|x}
\end{equation}

The prediction of interest $\p$ is also considered to be linearly dependent on $\x$ with a coefficient matrix $\Y$, and an additive error denoted by $\np$, such that $\p = \Y \x + \np$. In addition, it is  considered that the uncertainty in the value of the error is a zero mean Gaussian density with known covariance $\cov{\np}$. 
It is noted that this requires that $\nd$ and $\np$ are independent. If any correlation exists, it must be included in $\x$.
Finally, the conditional belief of the prediction given the system properties can now be defined as
\begin{equation}
p(\p | \x) \define N(\p; \Y\x, \cov{\np}).
\label{eq:pp|x}
\end{equation}
Note that if the system is believed to be well described by a deterministic model, the covariance of the prediction error could be considered negligible.

The calibration or inversion stage seeks to determine the posterior belief over the parameters $\x$ given the available data $\d$%
\vspace{-2mm}
\begin{equation*}
p(\x|\d) = \frac{p(\x) p(\d|\x)}{\int p(\x) p(\d|\x) d\x}.
\end{equation*}  
Here, the posterior density $p(\x|\d) = N(\x;\mean{\x|\d},\cov{\x|\d})$ is Gaussian with a mean and covariance given by
\begin{align}
\mean{\x|\d} & = \cov{\x} \G\t (\G \cov{\x} \G\t + \cov{\nd})^{-1} \d  ,
\label{eq:state-update} 
\\
\cov{\x|\d} & = 
\cov{\x} - \cov{\x} \G\t (\G \cov{\x} \G\t + \cov{\nd})^{-1} \G \cov{\x} .
\label{eq:statecov-update} 
\end{align}
The coefficient matrix of the data $\d$ in \cref{eq:state-update} is sometimes referred to as the optimal estimator, or gain matrix, and will be denoted by  $\E \define \cov{\x} \G\t (\G \cov{\x} \G\t + \cov{\nd})^{-1}$.

To propagate the posterior belief into the prediction of interest, requires the integration over the underlying parameters %
\begin{equation*}
p(\p|\d) = \int p(\p|\x) p(\x|\d) d\x.  
\end{equation*}
The prediction posterior density $p(\p|\d) = N(\p; \mean{\p|\d} , \cov{\p|\d})$ is Gaussian with mean and covariance given by
\begin{align*}
\mean{\p|\d} & = \Y \mean{\x|\d}  
\\
&= \Y \cov{\x} \G\t (\G \cov{\x} \G\t + \cov{\nd})^{-1} \d  ,
\\
\cov{\p|\d} &= \Y \cov{\x|\d} \Y\t + \cov{\np} 
\\
&= \Y\cov{\x} \Y\t + \cov{\np} 
- \Y \cov{\x} \G\t (\G \cov{\x} \G\t + \cov{\nd})^{-1} \G \cov{\x} \Y\t .   
\end{align*}

The overall calibration and prediction scheme is now summarized below. This sets the benchmark for the other schemes considered in this paper.

\begin{defn}[Optimal Scheme]
\label{defn:optimal-scheme} 
The optimal scheme is denoted by the set of linear Gaussian probability density functions $p(\x)$, $p(\d|\x)$ and $p(\p|\x)$, parameterized by the matrices $\cov{\x}$, $\G$, $\cov{\nd}$, $\Y$, and  $\cov{\np}$, as defined in \crefrange{eq:px}{eq:pp|x}.

For a given dataset $\d$, the posterior belief over the predictions of interest is the Gaussian density $p(\p|\d) = N(\p;\mean{\p|\d}, \cov{\p|\d})$, with mean and covariance defined by the functions $\predMean{}(\cdot)$ and $\predCov{}(\cdot)$ respectively
\begin{align}
\mean{\p|\d} &\define \predMean{}(\cov{\x},\G,\cov{\nd},\Y,\cov{\np},\d)
\label{eq:pred-update}
\\
&\define 
\Y \cov{\x} \G\t (\G \cov{\x} \G\t + \cov{\nd})^{-1} \d ,
\nonumber
\\
\cov{\p|\d} &\define \predCov{}(\cov{\x},\G,\cov{\nd},\Y,\cov{\np})
\label{eq:predcov-update}\\
&\define 
\Y\cov{\x} \Y\t + \cov{\np} 
- \Y \cov{\x} \G\t (\G \cov{\x} \G\t + \cov{\nd})^{-1} \G \cov{\x} \Y\t   .
\nonumber
\end{align}
Note that the covariance matrix of the posterior is not a function of the data $\d$.
\end{defn}\vspace{3pt}

\begin{figure}[tb]%
\small
\centering
\includegraphics[width=0.3\columnwidth]{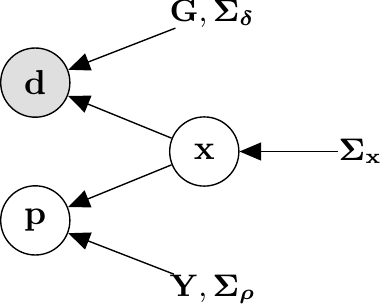}
\caption{Bayesian network depicting the structure of the densities $p(\x)$, $p(\d|\x)$, $p(\p|\x)$ that employ the high fidelity model.
}%
\label{fig:full-scheme-BN}%
\end{figure}

The structure of the above prediction scheme that encapsulates the prior densities $p(\x)$, $p(\d|\x)$, $p(\p|\x)$ that are based on the high fidelity model is depicted graphically in \cref{fig:full-scheme-BN}. %

\section{Model Simplifications}
\label{sec:model-simplifications}

The simplified computational model is considered to contain an approximate system simulator, with exposed parameters $\v$ and linear structure analogous to that defined for the high fidelity model
\begin{align}
\d &\approx \tG \v + \nd ,
\label{eq:simple-data-model}
\\
\p &\approx \tY \v + \np .
\label{eq:simple-predict-model}
\end{align}
Here $\tG$ and $\tY$ are linear data and prediction matrices of a simplified system simulator. The exact form of the approximations within these relationships are of critical interest and will be explored by explicitly representing the simplifications involved. 

To achieve this consider initially that the parameters $\v$ are somehow physically inspired and have some meaning to the modeller in describing aspects of the system. %
This allows the parameters of the simplified model to be used to describe a restricted version of the high fidelity model with parameters $\x'\in\Re^{D_{\x}}$ determined by the parameter vector of the simplified model $\v'\in\Re^{D_{\v}}$ via a matrix $\C \in \Re^{D_{\x}\times D_{\v}}$ 
\begin{equation}
\x' = \C \v'.
\label{eq:simplification-model}
\end{equation}
This explicit representation of model simplifications allows aspects of the system to be ignored \cite{doherty_short_2010, white_quantifying_2014} and allows for the incorporation of arbitrary linear parameterization schemes, such as spatial and temporal homogeneity \cite{mclaughlin_reassessment_1996}. 	

With this definition, the simplified data and prediction matrices may be  rewritten in terms of the matrices of the high fidelity model
\begin{align}
\tG & \define \G \C ,
\label{eq:simplifications-data}
\\
\tY & \define \Y \C .
\label{eq:simplifications-predictions}
\end{align}
The matrix $\C$ will be referred to as a simplification matrix and provides a way of linking the simplified simulator to the original high fidelity reference model of how the system functions.%

Finally, it is important to note that physical intuition is not strictly required when defining the simple model as for any $\G$, $\Y$ and $\tG$, $\tY$, 
such that $\tmatrix{\G\\ \Y}$ has full row rank, 
there is always a $\C$ that satisfies \cref{eq:simplifications-data} and \cref{eq:simplifications-predictions}. However, this may not be unique and thus there are potentially many different ways to interpret the meaning of the simple model. Here, it is considered that the modeller specifies the interpretation by e.g. specifying $\C$ directly. 

\subsection{Representing Unmodelled Complexity}

A simplification matrix $\C \in \Re^{D_\x \times D_\v}$, divides the space $\Re^{D_\x}$ into two perpendicular subspaces: $\range(\C)$ and $\cokernel(\C)$. The column space contains the set of parameter vectors that can be explicitly represented by a low dimensional vector. The second subspace $\cokernel(\C)$ contains vectors that includes some degree of complexity that cannot be represented by a low dimensional vector. 

Before defining these further, the following assumption is made.
\begin{assump} \label{assump:c-full-column-rank}
A simplification matrix $\C$ has full column rank, i.e. $\rank(\C)=D_\v$.
\end{assump}\vspace{3pt}
It is noted that if a simplification matrix $\C$ does not have full column rank, then there will be parameter combinations of the simple model that have the same influence on the high fidelity model, which implies that the simple model is not as simple as it could be for the same expressive power. 

Now, consider the singular value decomposition of $\C$ under the above assumption
\begin{equation}
\C = \U \S \V\t = \matrix{\Ucol{\C} & \D} \matrix{ \Sing{\C} \\ 0} \Vrow{\C}\t  = \Ucol{\C} \Sing{\C} \Vrow{\C}\t,
\label{eq:param-subspace}
\end{equation}
where $\Ucol{\C}$ is the collection of orthogonal unit vectors that span the column space of $\C$ and similarly $\D$ spans the subspace perpendicular to this, the cokernel of $\C$.

This enables the parameter vector of the high fidelity model to be expanded as
\begin{equation}
\x = \C\v + \D\u  ,
\label{eq:state-decomposition}
\end{equation} 
where $\v$ is the parameter vector of the simple model and $\u$ is a random vector that captures all the unmodelled complexity. Finally, it is noted that this expansion is unique and only dependent on the definition of the simplification matrix $\C$.

Under the above expansion, the data and prediction is related to the two components $\v$ and $\u$ through  
\begin{align}
\d &= \G\x + \nd %
= \tG\v + \nd + \overbrace{\G\D\u}^{\boldsymbol\eta}  ,
\label{eq:simple-data-model-error}
\\
\p &= \Y\x + \np %
= \tY\v + \np + \underbrace{\Y\D\u}_{\boldsymbol\epsilon}   .
\label{eq:simple-predict-model-error}
\end{align}
From these equations, it is noted that the unmodelled components of the system will be expressed in the data whenever $\G\D \neq 0$ and in the predictions whenever $\Y\D \neq 0$. These additional error terms are denoted as $\boldsymbol\eta$ and $\boldsymbol\epsilon$ respectively and play a very different role to the other error terms $\nd$ and $\np$ as they are correlated via $\u$. It is noted that the additional error $\boldsymbol\eta$ introduced by the model simplifications was explicitly considered in \citet{mclaughlin_reassessment_1996} and \citet{tarantola_inverse_2005} and was used to define a composite measurement error model for the sum $\nd + {\boldsymbol\eta}$. However, the effects of the simplifications on the prediction, including the explicit correlation between $\boldsymbol\eta$ and $\boldsymbol\epsilon$, was not addressed.

In an ideal world it could be argued that the objective of any modelling exercise should be to construct a model that is a simplified version of how the system is believed to function that captures the full complexity of the available data and predictions of interest. That is, which has a $\C$ such that $\boldsymbol\eta = \boldsymbol\epsilon = 0$. Such a simplification matrix always exists, and will be referred to as an optimal simplification. 
\begin{defn}[Optimal Simplification] \label{defn:optimal-simplification}
Let $\G$ and $\Y$ be data and predictions matrices for a high fidelity system model. Then a simplification matrix $\C$ is optimal if 
\begin{equation}
\G\D = 0 \quad \text{and} \quad \Y\D = 0,
\label{eq:optimal-simplification}
\end{equation}
where $\D$ spans the cokernel of $\C$, as defined in \cref{eq:param-subspace}.
\end{defn}\vspace{3pt}
It is noted that the above definition of optimal simplification differs from that of \cite{doherty_short_2010, doherty_use_2011, watson_parameter_2013} in that it explicitly considers the nature of the predictions.

An optimal simplification matrix, $\C^*$, can be found for any given data and prediction matrices $\G$ and $\Y$, by first taking a singular value decomposition of the matrix $\Z=\tmatrix{\G \\ \Y}$ 
\begin{equation}
\Z = \matrix{\G \\ \Y} = 
\matrix{\Ucol{\Z} & \Ulnu{\Z}} 
\matrix{ \Sing{\Z} & 0 \\ 0 & 0} 
\matrix{\Vrow{\Z}\t \\ \Vnul{\Z}\t}.
\end{equation}
Now, an optimal simplification matrix is given by $\C^* = \Vrow{\Z}$.

It is considered that an optimally simplified model is difficult, or even achievable, to obtain in practice and the remainder of the paper will focus on understanding and overcoming the issues caused by suboptimal simplifications.

\section{Naive Use of Simplified Models}
\label{sec:naive-use}

Here, a prediction scheme is defined that first infers the parameters of the simple model and then propagates them into the prediction, but ignores the potential for unmodelled complexity to be expressed in the data or predictions. This represents a standard probabilistic Bayesian approach to model calibration and prediction. It is shown that this scheme is generally not conservative as the uncertainty in the predictions is underestimated. 

\subsection{Prior Information}

The scheme will have the same basic structure as the optimal benchmark scheme. In particular it will include an explicit representation of a prior over the parameters $\ap\naive(\v)$, a data likelihood $\ap\naive(\d|\v)$ and conditional density for the prediction $\ap\naive(\p|\v)$. Here the superscript $\naivea$ denotes approximate distributions associated with this naive method.

The conditional densities $\ap\naive(\d|\v)$ and $\ap\naive(\p|\v)$ are defined directly with the simplified data and prediction matrices $\tG$ and $\tY$ respectively. Furthermore, the covariances are set to those used in the optimal scheme i.e.
\begin{align}
\ap\naive(\d|\v) &\define N(\d; \tG\v, \cov{\nd}),
\label{eq:pd|v}
\\
\ap\naive(\p|\v) &\define N(\p; \tY\v, \cov{\np}).
\label{eq:pp|v}
\end{align}
This formulation of the data and prediction densities is equivalent to ignoring the approximations in \cref{eq:simple-data-model} and \cref{eq:simple-predict-model} introduced by the model simplifications. 

Now, to be somewhat rigorous, the specification of a prior distribution on the parameters $\v$ is performed by explicitly considering the nature of the simplification and the uncertainty in the parameters of the high fidelity model. This is accomplished by considering a transformation that propagates the prior distribution over the parameters of the high fidelity model $\x$ into the joint space formed by the parameter vector of the simple model $\v$ and the vector $\u$ that denotes the unmodelled complexity. This is defined in the following proposition.

\begin{prop}[Uncertainty Propagation] \label{prop:uncertainty-propagation}
Let $p(\x) = N(\x;0;\cov{\x})$ be a zero mean Gaussian density, with covariance $\cov{\x}$, that captures the prior uncertainty in the parameters of a high fidelity model. Also, let $\C$ be a simplification matrix that describes a simplified model. 

Then, the joint probability density function for $\v$ and $\u$, representing the modelled and unmodelled components of the simple model respectively, is the zero mean correlated Gaussian density
\begin{equation*}
p(\v,\u) = N\Bigl( \matrix{\v \\ \u};\matrix{0 \\ 0},\matrix{\cov{\v} & \cov{\v\u} \\ \cov{\v\u}\t & \cov{\u}} \Bigr),
\end{equation*}
where the covariance terms are given by
\begin{equation*}
\cov{\v} = \C\pinv \cov{\x} \C\pinv\t, \; \cov{\u} = \D\t \cov{\x} \D, \; \text{and} \;\; \cov{\v\u} = \C\pinv \cov{\x} \D.
\end{equation*}
Here, $(\cdot)\pinv$ is the pseudoinverse operator and $\D$ spans the cokernel of $\C$ as defined in \cref{eq:param-subspace}.
\end{prop}\vspace{3pt}\noindent
See \Cref{sec:proof1} for a proof.

This proposition defines the covariance matrix $\cov{\v}$, that represents the  uncertainty in the parameters of the simplified model, to be the orthogonal projection of $\cov{\x}$ onto the column space of $\C$. This is the most rigorous approach to specifying the prior over the parameters of the simple model as it preserves the uncertainty that is believed to exist and is consistent with the physical intuition used to construct the simplified model. 

This marginal density is used to define the prior of the naive prediction scheme
\begin{equation*}
\ap\naive(\v) = N(\v ; 0, \cov{\v} ).
\end{equation*}

\subsection{Naive Calibration and Prediction}

The naive calibration and prediction scheme is now defined that generates a posterior via the standard rules of probability theory. This is similar to the optimal scheme, but without consideration to the simplifications within the model's data and prediction equations. 

\begin{defn}[Naive Scheme]
\label{defn:naive-scheme}
A naive prediction scheme is denoted by the simplified linear Gaussian  density functions $\ap\naive(\v)$, $\ap\naive(\d|\v)$ and $\ap\naive(\p|\v)$ parameterized by the matrices $\cov{\v}$, $\tG$, $\cov{\nd}$, $\tY$, and  $\cov{\np}$. 

Furthermore, the prior covariance matrix $\cov{\v}$ is defined in terms of a high fidelity model using \cref{prop:uncertainty-propagation}, that is
\begin{equation}
\cov{\v}  \define \C\pinv \cov{\x} \C\pinv{}\t,
\label{eq:prior-cov-simple}
\end{equation}
where $\cov{\x}$ denotes a covariance matrix that describes the uncertainty in the parameters of the high fidelity model, and $\C$ denotes the simplification matrix that links the parameters of the simple and high fidelity models.

For a given dataset $\d$, the posterior belief over the predictions of interest generated by this naive scheme is the Gaussian density $\ap\naive(\p|\d) = N(\p;\tmeannaivea{\p|\d}, \tcovnaivea{\p|\d})$, with mean and covariance defined as 
\begin{align}
\tmeannaivea{\p|\d} &\define \predMean{}(\cov{\v},\tG,\cov{\nd},\tY,\cov{\np},\d)  ,
\label{eq:naive-predict-mean}
\\ 
\tcovnaivea{\p|\d} &\define \predCov{}(\cov{\v},\tG,\cov{\nd},\tY,\cov{\np})  ,
\label{eq:naive-predict-cov}
\end{align}
where the functions $\predMean{}(\cdot)$ and $\predCov{}(\cdot)$ are given in \cref{defn:optimal-scheme}.
\end{defn}\vspace{3pt}

\subsection{Performance}

To assess the performance of the naive scheme the notion of conservativeness, discussed in \Cref{sec:problem-definition}, is first defined by considering the squared error in the mean prediction. This is performed for a pair of probability density functions in \cref{defn:conservative}, and extended to calibration and prediction schemes in \cref{defn:conservative-scheme}. A further generalization of these definitions, which explicitly incorporates the utility function of the subsequent decision problem, is proposed in \Cref{sec:conclusion}.

\begin{figure}[tb]%
\small
\centering
\includegraphics[width=0.7\columnwidth]{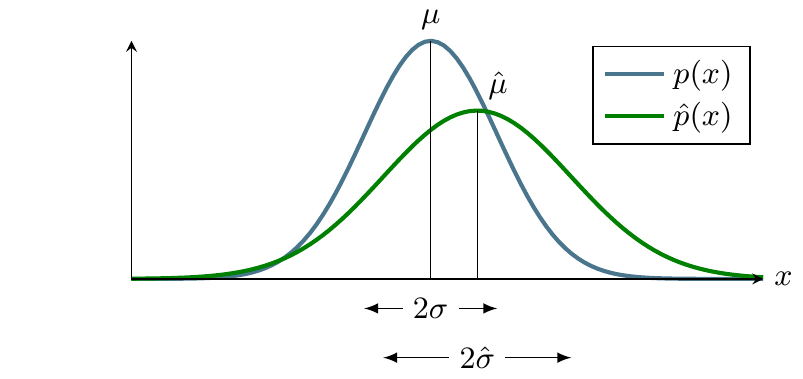}
\caption{The Gaussian probability density function $\ap(x)$, with mean $\hat\mu$ and standard deviation $\hat\sigma$, is a conservative estimate of the Gaussian density $p(x)$, with mean $\mu$ and standard deviation $\sigma$, as $\hat\sigma^2 \geq \sigma^2 + (\mu-\hat\mu)^2$.}
\label{fig:conservative}%
\end{figure}

\begin{defn}[Conservative Density] \label{defn:conservative}
Let $p(\p)$ and $\ap(\p)$ be density functions defined over the random vector $\p$. Furthermore, let $\tmean{\p}$ be the mean of $\ap(\p)$. Then $\ap(\p)$ is defined as a conservative approximation of the reference density $p(\p)$ if the approximate density's expected mean squared error $E_{\ap(\p)} \{ (\tmean{\p} - \p)(\tmean{\p}-\p) \t \}$, does not increase when the expectation is instead taken with respect to reference density $p(\p)$
\begin{equation}
E_{\ap(\p)} \{ (\tmean{\p} - \p)(\tmean{\p}-\p) \t \} \succeq E_{p(\p)} \{ (\tmean{\p} - \p)(\tmean{\p}-\p) \t \}.
\label{eq:conservative}
\end{equation}
Where $\A \succeq \B$ requires that $\A - \B$ is positive semi-definite. 
\end{defn}\vspace{3pt}
It is noted that condition \cref{eq:conservative} can be rewritten in terms of the means $\mean{\p}$,  $\tmean{\p}$ and covariances $\cov{\p}$, $\tcov{\p}$ of the two densities, and yields the condition
\begin{equation*}
\tcov{\p} \succeq \cov{\p} + (\tmean{\p}-\mean{\p})(\tmean{\p}-\mean{\p})\t.
\end{equation*}
An example of a conservative density, that satisfies \cref{defn:conservative}, is given in \cref{fig:conservative}.

The notion of conservativeness is now generalized to a prediction scheme. This is performed by considering an expectation over typical data. 

\begin{defn}[Conservative Scheme] \label{defn:conservative-scheme}
Let the densities $p(\x)$, $p(\d|\x)$, and $p(\p|\x)$ denote the prior information of a reference scheme. Also, let $p(\p|\d)$ denote the posterior density this scheme generates for a given dataset $\d$. Furthermore, let $\ap(\p|\d)$ denote a posterior density generated by an approximate scheme. 

For the given dataset $\d$, the degree of conservativeness of the approximate posterior is denoted by the function
\begin{equation}
\Omega(\d) = E_{\ap(\p|\d)} \{ (\tmean{\p|\d} - \p)(\tmean{\p|\d}-\p) \t \} - E_{p(\p|\d)} \{ (\tmean{\p|\d} - \p)(\tmean{\p|\d}-\p) \t \} .
\end{equation}
Now, the approximate scheme is defined as conservative if the expectation of the function $\Omega(\d)$ is positive semi-definite
\begin{equation}
 E_{p(\d)} \{ \Omega(\d) \} \succeq 0.
 \label{eq:conservative-scheme}
\end{equation}
Here, the density $p(\d)$ denotes how probable different datasets are under the prior knowledge of the reference scheme and is given by $p(\d)=\int p(\x)p(\d|\x)d\x$.
\end{defn}\vspace{3pt}
 
It is noted that for the linear Gaussian densities considered in this work, the covariance of the posterior distributions are independent of the data and \cref{eq:conservative-scheme} can be simplified as 
\begin{equation}
\tcov{\p|\d}\succeq \cov{\p|\d} + E_{p(\d)} \big\{  (\tmean{\p|\d}-\mean{\p|\d})(\tmean{\p|\d}-\mean{\p|\d})\t \big\}.
\end{equation}
Thus, a scheme is considered as conservative if it generates a posterior covariance matrix $\tcov{\p|\d}$ that is inflated with respect to $\cov{\p|\d}$ by an amount dependent on the average squared difference in the posterior means.

It is now of interest to consider the performance of the naive scheme and determine when it is conservative with respect to the optimal benchmark scheme. This is performed in the following proposition.

\begin{prop}[Performance of Naive Scheme] \label{prop:naive-performance}
Suppose the matrices $\cov{\x}$, $\G$, $\cov{\nd}$, $\Y$, $\cov{\np}$ define an optimal calibration and prediction scheme, with posterior density denoted by $p(\p|\d) = N(\p;\mean{\p|\d},\cov{\p|\d})$ for  arbitrary dataset $\d$. 

Let $\C$ denote a simplification matrix such that the matrices $\cov{\v}=\C\pinv \cov{\x} \C\pinv{}\t$, $\tG=\G\C$, $\cov{\nd}$, $\tY=\Y\C$, $\cov{\np}$ define a naive prediction scheme for a simplified model. 
Now, consider the posterior density generated by the naive scheme $\ap\naive(\p|\d) = N(\p; \tmeannaivea{\p|\d}, \tcovnaivea{\p|\d})$.
\begin{enumerate}%
\item If $\C$ is an optimal simplification, then the generated mean and covariance of the naive prediction scheme are equivalent to those of the optimal prediction scheme, that is
\begin{equation*}
\tmeannaivea{\p|\d} = \mean{\p|\d}  \quad \text{and} \quad \tcovnaivea{\p|\d} = \cov{\p|\d} \quad \text{for all $\d$}.
\end{equation*}

\item If $\C$ is a suboptimal simplification, then the naive scheme is conservative, provided the following condition holds
\begin{equation}
\Y\D = \tY \E\naive\G\D,
\label{eq:naive-balanced-error}
\end{equation}
where $\E\naive \define \cov{\v} \tG\t (\tG \cov{\v} \tG\t + \cov{\nd})^{-1}$ denotes the estimator matrix of the naive scheme and $\D$ spans the cokernel of $\C$ as defined in \cref{eq:param-subspace}.

\item \label{XXXX}
If $\C$ is a suboptimal simplification, condition \cref{eq:naive-balanced-error} does not hold, $\u$ and $\v$ are considered independent, and the covariance in the unmodelled complexity is non-zero $\cov{\u} \neq 0$, then the naive scheme is strictly non-conservative.

\end{enumerate}

\end{prop}\vspace{3pt}\noindent
See \Cref{sec:proof2} for a proof.

Under a suboptimal simplification, the non-conservative nature of the prediction scheme is caused by the influence of unmodelled complexity on the data and/or predictions. If it influences the predictions, then there is a direct bias effect. If it influences the data, then this causes the estimated parameters of the simple model to become biased by forcing them to take on surrogate roles to compensate for the simplifications. Furthermore, this bias in the estimated parameters is propagated to the prediction. Neither of these effects are taken into account by the naive prediction scheme and the uncertainty is underestimated.

The condition in \cref{eq:naive-balanced-error}, which guarantees the scheme is conservative, requires that these two sources of errors in the prediction caused by the unmodelled complexity to exactly cancel each other out. This is unlikely to hold in practical scenarios without careful attention to the data, predictions, simplifications, and the type of prior knowledge available. This condition will form the basis of the data driven prediction scheme introduced in \Cref{sec:overcomming-simplifications}.

Finally, it is noted that for the general case when $\v$ and $\u$ are not independent and \cref{eq:naive-balanced-error} does not hold, the naive scheme is still not guaranteed to be conservative. However there are more special cases which may  produce conservative posterior densities. The conditions that delineate the strictly conservative from the non-conservative scenarios are likely to be of limited interest in realistic problems and have not been enumerate here.

\subsection{Summary}

This section has formally defined a probabilistic calibration and prediction scheme that embeds the standard separation of modelling, calibration and prediction. In particular, physical knowledge is used to define a prior probability distribution for the parameters included in the simplified model. Also, the conditional data and prediction probability distributions are defined by ignoring the simplifying assumptions used to construct the data and prediction equations. Furthermore, it is shown that a true characterization of predictive uncertainty typically requires an optimally simplified model. For suboptimally simplified models it is shown that this scheme generally underestimates the uncertainty in the predictions and is not conservative. 

\begin{figure}[tb]%
\centering
\includegraphics[width=0.35\columnwidth]{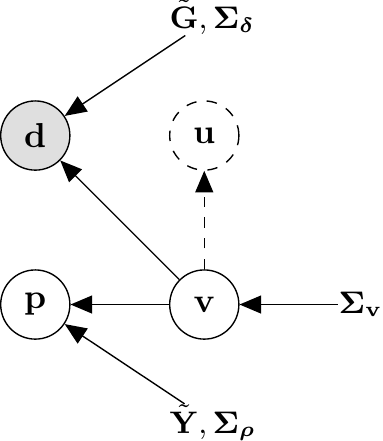}
\caption{Bayesian network depicting the structure of the densities $\ap\naive(\v)$, $\ap\naive(\d|\v)$, and $\ap\naive(\p|\v)$ employed by the naive scheme. Also displayed in dashed lines is the dependency on unmodelled complexity $\u$ under the conditions of an optimal simplification. Note that under an optimal simplification, the unmodelled complexity does not have any direct effect on the data or predictions.}%
\label{fig:naive-scheme-BN}%
\end{figure}

The dependency structure of the probability distributions for this naive scheme is depicted in \cref{fig:naive-scheme-BN}. Also displayed is the dependency on the unmodelled complexity for an optimally simplified model. It is noted that under this condition the data and predictions are conditionally independent of the unmodelled complexity $\u$ given the model parameters $\v$.

Finally, it is noted that the non-conservative, or overconfident, nature of predictions generated by naively applying probabilistic Bayesian methods on simplified models is not a new finding. In particular, the result can be considered as a generalization of \cite{doherty_short_2010, doherty_use_2011, white_quantifying_2014} and is consistent with arguments of \citet{beven_equifinality_2001}.

\section{Overcoming Model Simplifications}
\label{sec:overcomming-simplifications}

It is of interest now to understand what to do about the issue of overconfidence when a suboptimal model is used. Firstly, it is noted that this issue may not be a problem. For instance, the prediction uncertainty may turn out to be larger than expected, and this in itself may provide sufficient information to allow a given management decision to be made. %
However, the situation becomes more difficult when it is necessary to determine an accurate or at least conservative prediction probability distribution, for instance as required by \citet{doherty_groundwater_2013}. In this scenario the modeller may opt to:
\begin{itemize}

\item Expand the sources of uncertainty considered in the model through less simplistic modelling assumptions with the hope of producing an optimally simplified model. This is not just about representing greater spatial/temporal detail in the model \cite{hunt_are_2007}, but more importantly the goal should be the inclusion of more sources of uncertainty. As an example, if there exists uncertainty over the presence of some important structural feature, a multi-hypothesis, or multi-model, analysis may be needed \cite{poeter_all_2007, poeter_mma_2007, rojas_application_2010}. 

\item Directly model the additional errors introduced by the model simplifications. For instance by defining an additional probabilistic model for $p(\boldsymbol{\eta},\boldsymbol{\epsilon})$ introduced in \cref{eq:simple-data-model-error} and \cref{eq:simple-predict-model-error} that captures the uncertainty the modeller has in how well the simplistic model represents the behavior of the system. Such approaches have been  developed in \cite{kennedy_bayesian_2001, craig_bayesian_2001, goldstein_probabilistic_2004, goldstein_bayes_2006, rougier_probabilistic_2007, rougier_uncertainty_2014}. 

\item Within the context of a dynamical system, remove the deterministic assumption on how the system evolves over time. This allows the simplifications to be represented by an error model within the modelled transition function. Such a scheme is employed by \citet{vrugt_improved_2005}. Alternatively, the assumption that the model parameters are time invariant can be relaxed such that the parameters can change over time to better match the underlying system behavior and inject greater uncertainty in the predictions \cite{reichert_analyzing_2009}. 

\item Modify the way that the data is used through changing the likelihood model $p(\d|\v)$ such that less information is propagated into the prediction via the model parameters. 
The generalized likelihood uncertainty estimation method \cite{beven_equifinality_2001} can be considered as an example of this approach.

\end{itemize}

In the remainder of this section, two additional approaches are defined, along with explicit conditions that determine when they are appropriate. Firstly a scheme is defined that allows the posterior of the ideal prediction scheme to be reproduced using a suboptimally simplified model through the appropriate modifications of the prior covariances. Secondly, structural considerations of the data and predictions are considered and an approach developed that allows the simplifications to be overcome through the use of the right calibration data.

\subsection{Optimal Use of Simplified Models}

The naive scheme developed previously was not conservative for a suboptimally simplified model. Here a prediction scheme is defined that is capable of reproducing the optimal scheme, even under suboptimal simplifications. 

To start, consider the probability densities $\ap\opt(\v)$, $\ap\opt(\d|\v)$, $\ap\opt(\p|\v)$ to be defined in a similar manner to the naive scheme, but parameterized in terms of new covariance matrices $\tcovopta{\v}$, $\tcovopta{\nd}$ and $\tcovopta{\np}$ such that 
\begin{align*}
\ap\opt(\v) &= N(\v;0,\tcovopta{\v}) ,
\\
\ap\opt(\d|\v) &= N(\d;\tG \v,\tcovopta{\nd}) ,
\\
\ap\opt(\p|\v) &= N(\p;\tY \v,\tcovopta{\np}) .
\end{align*}
Here the covariance matrices will be considered as adjustable such that they can be chosen in such a way that they compensate for the simplifications and allow the optimal posterior to be reproduced. Thus, it is of interest to select $\tcovopta{\v}$, $\tcovopta{\nd}$ and $\tcovopta{\np}$ such that the posterior density that is produced by their combination, $\ap\opt(\p|\d)$, replicates the posterior of the optimal scheme, i.e.
\begin{equation}
\ap\opt(\p|\d) = p(\p|\d) \quad \text{for all $\p$, $\d$.}
\label{eq:optimal-posterior-equality}
\end{equation}
The posterior density $\ap\opt(\p|\d)$ is Gaussian, with mean and covariance defined in a similar fashion to the optimal scheme. Thus, the above condition will be satisfied when the means and covariances are equivalent, which occurs when
\begin{subequations}
\label{eq:optimal-inference-conditions}
\begin{align}
\tY\tcovopta{\v}\tG\t[ \tG\tcovopta{\v}\tG\t  + \tcovopta{\nd}]\i 
&= \Y\cov{\x}\G\t[ \G\cov{\x}\G\t  + \cov{\nd}]\i 
\end{align}
and
\begin{multline}
\tY \tcovopta{\v} \tY\t + \tcovopta{\np} - \tY\tcovopta{\v}\tG\t[ \tG\tcovopta{\v}\tG\t  + \tcovopta{\nd}]\i  \tG \tcovopta{\v} \tY\t 
= 
\\
\Y\cov{\x}\Y\t + \cov{\np}- \Y\cov{\x}\G\t[ \G\cov{\x}\G\t  + \cov{\nd}]\i  \G\cov{\x}\Y\t .
\end{multline}
\end{subequations}
If these conditions can be met, it defines a set of optimal covariance matrices of the simplified probability density functions such that when they are used to generate a posterior, the performance of the optimal scheme is replicated. 

\subsubsection{Highly Parameterized Models}

The above conditions will now be specialized for a highly parameterized model, where the calibration problem is under-constrained. Such models are recommended by \citet{hunt_are_2007} and \citet{white_quantifying_2014}, and the results will explicit determine when such approaches are appropriate. This will be performed by focusing attention only on the prior covariance matrix for the parameters of the simple model, such that $\tcovopta{\nd}$ and $\tcovopta{\np}$ remain the same as those used in the naive and optimal schemes, i.e. $\tcovopta{\nd}=\cov{\nd}$ and $\tcovopta{\np}=\cov{\np}$. Such a scheme is now defined.

\begin{defn}[Optimally Compensated Scheme]
\label{defn:opt-comp-scheme}
An optimally compensated scheme is denoted by the simplified linear Gaussian probability density functions $\ap\opt(\v)$, $\ap\opt(\d|\v)$ and $\ap\opt(\p|\v)$ parameterized by the matrices $\tcovopta{\v}$, $\tG$, $\cov{\nd}$, $\tY$, and $\cov{\np}$. 

To fully define the prior covariance matrix $\tcovopta{\v}$, let $\C$ denote a simplification matrix that links the simplified model to a high fidelity model with data and prediction matrices denoted by $\G$ and $\Y$ such that $\tY=\Y\C$ and $\tG=\G\C$. Furthermore, let $\cov{\x}$ denote the covariance matrix that describes the uncertainty in the parameters of this high fidelity model. The prior covariance matrix $\tcovopta{\v}$ is now defined as
\begin{equation}
\tcovopta{\v} =  \R\cov{\x}\R\t,
\label{eq:optimal-inference-conditions-output-driven}
\end{equation}
where $\R = \tZ\pinv \Z$, and $\Z=\tmatrix{\G \\ \Y}$, $\tZ=\tmatrix{\tG \\ \tY}$.

For a given dataset $\d$, the posterior belief over the predictions of interest generated by this scheme is the Gaussian density $\ap\opt(\p|\d) = N(\p;\tmeanopta{\p|\d}, \tcovopta{\p|\d})$, with mean and covariance defined as 
\begin{align}
\tmeanopta{\p|\d} &\define \predMean{}(\tcovopta{\v},\tG,\cov{\nd},\tY,\cov{\np},\d) ,
\label{eq:opt-predict-mean}
\\ 
\tcovopta{\p|\d} &\define \predCov{}(\tcovopta{\v},\tG,\cov{\nd},\tY,\cov{\np}) ,
\label{eq:opt-predict-cov}
\end{align}
where the functions $\predMean{}(\cdot)$ and $\predCov{}(\cdot)$ are given in \cref{defn:optimal-scheme}.
\end{defn}\vspace{3pt}

This scheme is now shown to be equivalent to the optimal scheme when applied to a suboptimal but highly parameterized model where the number of free parameters is at least as large as the number of linearly independent measurements and predictions of interest.

\begin{prop}[Performance of Compensated Scheme]
\label{prop:opt-comp-scheme}
Suppose the matrices $\cov{\x}$, $\G$, $\cov{\nd}$, $\Y$, $\cov{\np}$ define an optimal calibration and prediction scheme, with posterior density denoted by $p(\p|\d) = N(\p;\mean{\p|\d},\cov{\p|\d})$ for  arbitrary dataset $\d$. 

Let $\C$ denote a suboptimal simplification matrix and let the matrices $\tcovopta{\v} =  \R\cov{\x}\R\t$, $\tG=\G\C$, $\cov{\nd}$, $\tY=\Y\C$, and $\cov{\np}$ define an optimally compensated scheme, where $\R = \tZ\pinv \Z$ and $\Z=\tmatrix{\G \\ \Y}$, $\tZ=\Z\C = \tmatrix{\tG \\ \tY}$.
Now, consider the posterior density generated by the optimally compensated scheme $\ap\opt(\p|\d) = N(\p; \tmeanopta{\p|\d}, \tcovopta{\p|\d})$. 
If the simplification is chosen such that
\begin{equation}
\rank(\Z\C) = \rank(\Z)  .
\label{eq:rank-opt-cond}
\end{equation}
Then, the posterior mean and covariance generated by the optimally compensated scheme are equivalent to those of the optimal scheme 
\begin{equation*}
\tmeanopta{\p|\d} = \mean{\p|\d}  \quad \text{and} \quad \tcovopta{\p|\d} = \cov{\p|\d} \quad \text{for all $\d$}.
\label{eq:compensated-optimal}
\end{equation*}
\end{prop}\vspace{3pt}\noindent
See \Cref{sec:proof3} for a proof.

This demonstrates that given a model simplified in a suboptimal fashion, the optimal performance can be recovered through the adjustment of the assumed prior uncertainties. The main rank condition in \cref{eq:rank-opt-cond} is fairly easy to achieve in practice and is met when, e.g. the number of parameters is not smaller than the number of linearly independent measurements and predictions, and the simplified model matrix $\tZ=\Z\C$ has full (row) rank. 

A model may be highly parameterized due to the degrees of spatial variability in e.g. the modelled material properties that is included in the model (as in \cite{white_quantifying_2014}). However, the above result can apply to models of dynamical systems that use stochastic transition models (sometimes referred to as data assimilation methods). In these models the additional dynamic error terms that are incorporated at each time increment can be similarly viewed as a set of additional model parameters. 

\begin{figure}[tb]%
\centering
\includegraphics[width=0.35\columnwidth]{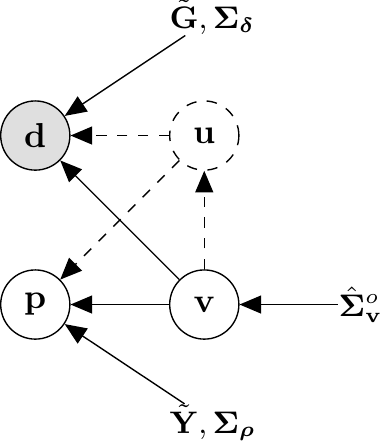}
\caption{Bayesian network depicting the structure of the densities $\ap\opt(\v)$, $\ap\opt(\d|\v)$, and $\ap\opt(\p|\v)$ employed by the optimally compensated scheme. Also displayed in dashed lines are the dependency on unmodelled complexity $\u$ under the conditions of a suboptimal simplification. Note that under such a simplification, the unmodelled complexity has a direct effect on the data and/or predictions.}%
\label{fig:optimal-scheme-BN}%
\end{figure}

The general structure of the optimally compensated calibration and prediction scheme is depicted with a Bayesian network in \cref{fig:optimal-scheme-BN}. It is noted that the difference between this scheme and the naive scheme lies only in how the prior covariance matrix for the parameters of the simple model is specified. They are both defined as a transformation of the covariance matrix $\cov{\x}$ that captures the uncertainty in the parameters for the high fidelity model. Recall that the naive scheme uses
\begin{equation*}
\cov{\v} =  \C\pinv\cov{\x}\C\pinv\t,
\end{equation*}
while the optimally compensated scheme uses
\begin{equation*}
\tcovopta{\v} =  \R\cov{\x}\R\t.
\end{equation*}
The matrix $\R$ is dependent on the simplification matrix $\C$ but it is also dependent on the data and predictions matrices $\G$ and $\Y$ of the high fidelity models. It is this explicit dependency on the data and predictions that allows the prior covariance matrix $\tcovopta{\v}$ to compensate for the errors in the data and prediction equations of the simplified model.

To gain additional intuition as to how the compensating prior covariance matrix $\tcovopta{\v}$ is related to $\cov{\v} $ used by the naive scheme, consider for the moment that $\v$ and $\u$ are independent. 
This enables the prior covariance matrix for the parameters of the high fidelity model to be decomposed as 
$\cov{\x} =  \C\cov{\v}\C\t + \D\cov{\u}\D\t$.
With this, the prior matrix $\tcovopta{\v}$ can be rewritten as
\begin{equation}
\tcovopta{\v} =  \cov{\v} + \cov{+},
\end{equation}
where $\cov{+} = \tZ\pinv\Z \D\cov{\u}\D\t  \Z\t\tZ\pinv{}\t$ and is positive semi-definite matrix. This means that for $\tcovopta{\v}$ to compensate for the simplifications, the parameters must be allowed greater flexibility than would be given by the naive use of $\cov{\v}$ as $\tcovopta{\v} \succeq  \cov{\v}$. 

\subsubsection{Weighted Least Squares Inversion}

In addition to the Bayesian arguments provided above, now consider a regularized weighted least squares formulation commonly used for calibration and inversion problems \cite{mclaughlin_reassessment_1996, tarantola_inverse_2005}. 
With the assumption of linear models, the mean vector $\tmeanopta{\v|\d}$ is also the maximum a posteriori estimate of the posterior parameter distribution of the simplified model $\ap\opt(\v|\d)\propto \ap\opt(\d|\v) \ap\opt(\v)$, and may be equivalently defined as the solution to the regularized weighted least squares optimization problem 
\begin{equation}
\tmeanopta{\v|\d} \define 
\argmin_{\v} (\d - \tG \v)\t \cov{\nd}\i (\d - \tG \v)  + \v\t \tcovopta{\v} {}\i \v  .
\end{equation}
Note that this explicitly employs the simplified forward model $\tG \v$.  Furthermore, the prediction generated by this estimate $\tmeanopta{\p|\d} = \tY \tmeanopta{\v|\d}$ is the mean and mode of the prediction posterior $\ap\opt(\p|\d)$. Due to the optimality of the compensated scheme, this point prediction is also the optimal minimum error variance prediction that the high fidelity model would generate.  

This demonstrates that a simplified, but highly parameterized model, can be calibrated through the careful selection of a Tikhonov regularizer to generate the optimal prediction.  Additionally, this optimal regularizer can also be used to characterize the predictive uncertainty of this point prediction through the use of the covariance formula in \cref{eq:opt-predict-cov}.

\subsubsection{Summary}

The introduced optimally compensated prediction scheme allows suboptimal model simplifications to be overcome through modifying the prior probability distribution for the parameters of the simplified model. This explicitly links: (i) the modelling problem, through the selection of a simplification matrix $\C$; (ii) the nature of the available data via $\G$; (iii) the predictions of interest defined by $\Y$; and (iv) how the simple model is calibrated, as the required regularization term $\tcovopta{\v}$ is explicitly  dependent on all three terms.

It is noted however, that this explicit dependency on the high fidelity model will make the generation of $\tcovopta{\v}$ problematic in practice as the matrix $\G$ and $\Y$ will not generally be available. Nevertheless, it formally defines how a prior regularization term for a suboptimal model should be selected.  

\subsection{Data Driven Predictions}
\label{sec:data-driven-predictions}

It was noted that the main barrier to applying the optimally compensated scheme defined above is the difficulty in determining the covariance matrix $\tcovopta{\v}$ that compensates for the simplifications. Here a new scheme is proposed that overcomes this issue by moving away from optimality and focusing on conservativeness. 

In particular, a specific class of predictions will be considered that allow the simplifications to be hidden behind the data. Thus, the objective is to examine how model simplifications can be overcome through the collection of the \emph{right data}. The intuition here is that data of a similar type to the predictions is often available, for example groundwater head measurements are often available when it is of interest to predict heads, similarly stream flow measurements are often available when predictions of stream flows are of interest, etc. 

\begin{figure}[tb]%
\centering
\includegraphics[width=0.4\columnwidth]{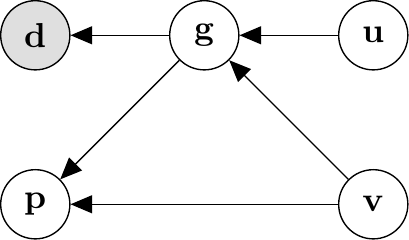}
\caption{Dependency structure of a prediction $\p$ where the unmodelled complexity $\u$ is hidden behind the data. Here the node $\g$ denotes an uncorrupted version of the data and is given by the deterministic relationship $\g =\G\x=\tG\v + \G\D\u$.}%
\label{fig:predict-class-BN-1}%
\end{figure}

The class of predictions that will be considered here incorporates two important principles: 
\begin{itemize}
\item The effects of all unmodelled components of the system on the predictions are captured by the data.
\item Any component of the system that effects the predictions, and is not captured by  the data, is explicitly represented in the simplified model. 
\end{itemize}
From these principles it is considered that the prediction of interest can be explicitly decomposed into an intermediate data term that is dependent on $\g=\G\x$ and a term wholly dependent on the parameters of the simple model, that is 
\begin{align}
\p &= \bar{\A}\g + \bar{\B}\v + \np ,
\label{eq:prediction-class}
\end{align}
where $\bar{\A}$ and $\bar{\B}$ are arbitrary matrices. Predictions with this structure are conditionally independent of $\u$ given $\g$ and $\v$, i.e. $p(\p|\g,\v,\u) = p(\p|\g,\v)$. This structure is represented by the Bayesian network in \cref{fig:predict-class-BN-1} and requires the prediction matrix to have the following form
\begin{equation}
\Y = \bar{\A}\G + \bar{\B}\C\pinv.
\label{eq:prediction-matrix-class}
\end{equation}
It will be shown in the sequel that even with this restricted form it is not possible to overcome the simplifications and further constraints must be imposed.

\subsubsection{Naive Scheme}

Now, consider the naive calibration and prediction scheme introduced in \cref{defn:naive-scheme}. Recall that this method will be conservative for a suboptimal simplification when condition \cref{eq:naive-balanced-error}, introduced in \cref{prop:naive-performance}, is satisfied. Specifically, this requires
\begin{equation*}
\Y\D - \tY \E\naive \G\D = 0  ,
\end{equation*}
where $\E\naive \define \cov{\v} \tG\t (\tG \cov{\v} \tG\t + \cov{\nd})^{-1}$.
For a prediction matrix $\Y$ with structure consistent with  \cref{eq:prediction-matrix-class}, this condition can be rewritten as
\begin{multline*}
\Y\D - \tY \E\naive \G\D   = \bar{\A}[\I - \tG\cov{\v}\tG\t [\tG\cov{\v}\tG\t+\cov{\nd}]\i] \G \D  
\\
 - \bar{\B} \cov{\v}\tG\t [\tG\cov{\v}\tG\t+\cov{\nd}]\i \D = 0 .
\end{multline*}
Now, under non-trivial conditions, this holds when
\begin{subequations}
\label{eq:nB-conditions}
\begin{equation}
\bar{\B} \cov{\v}\tG\t  = 0 ,
\label{eq:nB-conditions-1}
\end{equation}
and 
\begin{equation}
\tG\cov{\v}\tG\t [\tG\cov{\v}\tG\t+\cov{\nd}]\i = \I.
\label{eq:nB-conditions-2}
\end{equation}
\end{subequations}
Condition \cref{eq:nB-conditions-1} requires the random vector $\bar{\b}=\bar{\B}\v$ to be uncorrelated with $\tg = \tG\v$ under the prior covariance $\cov{\v}$. 
Furthermore, condition \cref{eq:nB-conditions-2} requires $\tG\cov{\v}\tG\t \gg \cov{\nd}$ and occurs when the data is perfect such that $\cov{\nd} \approx 0$, or when the prior knowledge in the subspace that is informed by the data (i.e. $\rowspace(\tG)$) is very weak such that $\tG\cov{\v}\tG\t \approx \infty $. 

These conditions ensure that: (i) the subset of parameters of the simple model that directly influence the predictions, represented by $\bar{\b}=\bar{\B}\v$, cannot be estimated from the data and (ii) the simple model can exactly reproduce the data.

It is important to note however that these conditions are unlikely to hold in practical scenarios and the naive scheme is not guaranteed to be conservative, even for the predictions with the structure considered in \cref{eq:prediction-class}. 

\subsubsection{Data Driven Scheme}

To overcome the above issues, it is proposed to ensure conservativeness through the judicial removal of information such that the conditions defined in \cref{eq:nB-conditions} can be satisfied. This will occur in two main areas: 
\begin{itemize}
	\item An easily computable inflation of the prior covariance matrix $\cov{\v}$.
	\item The application of a preprocessing or filtering step that may discard or combine some components of the data.
\end{itemize}
Additionally, the class of predictions must be restricted further than those allowed by \cref{eq:prediction-class}.

To start, let $\F \in \Re^{D_{\d'} \times D_{\d}}$ denote a data filtering matrix such that 
\begin{equation}
\d' \define \F \d ,
\label{eq:data-filter}
\end{equation}
where $D_{\d'} \leq D_{\d}$. Similarly define the transformed data matrices as $\G' \define \F\G$, and $\tG' \define \F\tG$, and the transformed data error covariance as $\cov{\nd}' \define \F\cov{\nd}\F\t$.  The purpose of the filter $\F$ will be elaborated on later. Note however,  filtering is optional and the identity matrix can be used $\F = \I$.

Now, consider the singular value decomposition of $\tG'$
\begin{equation}
\tG' \define \F\tG \define 
\matrix{\U_1 & \U_2} 
\matrix{\S_1 & 0 \\ 0 & 0} 
\matrix{\V_1\t \\ \V_2\t}
= \U_1 \S_1 \V_1\t.
\label{eq:tG-svd}
\end{equation}
The last expression $\U_1 \S_1 \V_1\t$ only includes the nonzero singular values, and will be referred to as the compact SVD. 
With this expansion, the vectors $\v_1 \define \V_1\t \v$ and $\v_2 \define \V_2\t \v$ define the rowspace and nullspace components of the parameters of the simple model for the given filtered data matrix $\tG' = \F\tG$. To meet condition \cref{eq:nB-conditions-2}, the prior covariance will be inflated such that all information pertaining to the rowspace of $\tG'$, i.e. the subspace spanned by $\V_1$, is removed. In addition any correlation in the random vectors $\v_1$ and $\v_2$ will be ignored. Such a calibration and prediction scheme that embodies these properties is now defined.

\begin{defn}[Data Driven Scheme]
\label{defn:data-scheme}
For a given filtering matrix $\F\in\Re^{D_{\d'}\times D_{\d}}$ with $D_{\d'} \leq D_{\d}$, a data driven scheme is denoted by the set of simplified linear Gaussian probability density functions $\ap\dat(\v)$, $\ap\dat(\d'|\v)$ and $\ap\dat(\p|\v)$ parameterized by the matrices $\tcovdata{\v}$, $\tG'=\F\tG$, $\cov{\nd}'=\F\cov{\nd}\F\t$, $\tY$, and $\cov{\np}$.

To fully define the prior covariance matrix $\tcovdata{\v}$, let the covariance matrix $\cov{\v}$ denote the uncertainty in the parameters of the simplified model defined in \cref{prop:uncertainty-propagation} as $\cov{\v}  \define \C\pinv \cov{\x} \C\pinv{}\t$.
Furthermore, let the matrices $\V_1$ and $\V_2$ each denote a set of orthonormal column vectors that form a basis for the rowspace and nullspace of $\tG'$, e.g. as defined in \cref{eq:tG-svd}. 
The prior covariance matrix $\tcovdata{\v}$ is now defined by the limit
\begin{equation}
\tcovdata{\v} \define \lim_{\alpha \rightarrow \infty}
\alpha \V_1 \V_1\t +  \V_2\V_2\t\cov{\v}\V_2\V_2\t.
\label{eq:data-driven-inflation}
\end{equation}

For a given dataset $\d$ the posterior belief over the predictions of interest generated by the data driven scheme is the Gaussian density $\ap\dat(\p|\d) = N(\p;\tmeandata{\p|\d},\tcovdata{\p|\d})$, with mean and covariance given by 
\begin{align*}
\tmeandata{\p|\d} &\define \predMean{}(\tcovdata{\v},\tG',\cov{\nd}',\tY,\cov{\np},\d)  ,
\\ 
\tcovdata{\p|\d} &\define \predCov{}(\tcovdata{\v},\tG',\cov{\nd}',\tY,\cov{\np})  ,
\end{align*}
where the functions $\predMean{}(\cdot)$ and $\predCov{}(\cdot)$ are given in \cref{defn:optimal-scheme}.
\end{defn}\vspace{3pt}

\begin{figure}[tb]%
\linespread{1}
\centering
\includegraphics[width=0.4\columnwidth]{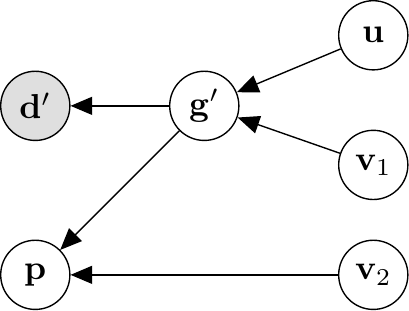}
\caption{Dependency structure of a prediction $\p$ where the unmodelled complexity $\u$ is hidden behind the data. Any induced parameter surrogacy in the parameters $\v_1$ cannot contaminate the predictions. Here the node $\g'$ denotes an uncorrupted version of the filtered data and is given by the deterministic relationship $\g' =\F\G\x=\F\tG\V_1\v_1 + \F\G\D\u$.}%
\label{fig:predict-class-BN-2}%
\end{figure}

Finally, a more specific class of predictions is considered than that defined in \cref{eq:prediction-matrix-class}, such that condition  \cref{eq:nB-conditions-1} will hold by construction. Specifically, the prediction must depend on the uncorrupted version of the filtered data, denoted by $\g' = \F\G\x$, or on the components of the parameters of the simple model that line in the nullspace of $\tG'$, denoted by $\v_2 = \V_2\t \v$. The prediction cannot depend on any rowspace components directly. Predictions with this structure can be written in the form
\begin{align}
\p &= \A\g' + \B\v_2 + \np ,
\label{eq:prediction-class-restrict}
\end{align}
where $\A$ and $\B$ are arbitrary matrices. Furthermore, the dependency structure of this restricted class of predictions is depicted in \cref{fig:predict-class-BN-2}. The key difference between this and the structure depicted in \cref{fig:predict-class-BN-1} is the explicit separation of $\v$ into $\v_1$ and $\v_2$ and the restriction that $\v_1$ cannot have a direct dependency on the prediction. That is, the prediction $\p$ is conditionally independent of $\u$ and $\v_1$ given $\g'$ and $\v_2$, or equivalently $p(\p|\g',\v_1,\v_2,\u)=p(\p|\g',\v_2)$. A prediction with this restricted structure has a prediction matrix of the form
\begin{equation*}
\Y = \A\F\G + \B \V_2\t \C\pinv.
\end{equation*}
It is noted that the matrices $\A$ and $\B$ are arbitrary, the important characteristic is the structure that links the data and predictions to the simplification.
This structure not only ensures that the  simplifications are hidden behind the data, but also that any surrogate roles that the parameter vector $\v_1$ is forced to undertake during calibration cannot adversely affect the predictions. 

It is now demonstrated that the data driven scheme is conservative for predictions that have this special structure.

\begin{prop}[Performance of Data Driven Scheme]
\label{prop:data-driven-performance}
Suppose the matrices $\cov{\x}$, $\G$, $\cov{\nd}$, $\Y$, $\cov{\np}$ define an optimal scheme with posterior denoted by $p(\p|\d) = N(\p; \mean{\p|\d}, \cov{\p|\d})$  for arbitrary dataset $\d$. 

Furthermore, let $\C$ denote a suboptimal simplification matrix and let the matrices $\F$, $\tcovdata{\v}$, $\tG'=\F\G\C$, $\cov{\nd}'=\F\cov{\nd}\F\t$, $\tY=\Y\C$, and $\cov{\np}$ denote a data driven scheme, where $\tcovdata{\v}$ is as specified in \cref{defn:data-scheme}. Additionally, let $\tmeandata{\p|\d}$ and $\tcovdata{\p|\d}$ denote the mean and covariance of the posterior produced by this scheme for the dataset $\d$. 

\begin{enumerate}%
\item If the filtering matrix $\F$ is selected such that $\tG'=\F\tG$ has full row rank, then the estimator matrix $\E\dat$ of the data driven scheme is equivalent to the pseudoinverse of $\tG'$ and can be expressed in terms of the compact SVD
\begin{equation}
\E\dat = (\tG')\pinv = \V_1 \S_1 \i \U_1\t.
\end{equation}
In addition, the posterior mean and covariance can be rewritten as
\begin{align*}
\tmeandata{\p|\d} 
&= \tY \E\dat \d,
\\ 
\tcovdata{\p|\d} 
&=  \tY \E\dat \cov{\nd} \E\dat {}\t \tY\t 
+ \tY \W \cov{\v} \W \tY\t   + \cov{\np} ,
\end{align*}
where $\W=\I - \V_1 \V_1\t$.

\item  If $\tG'=\F\tG$ has full row rank, and the high fidelity prediction matrix has the form
\begin{equation}
\Y = \A\F\G + \B \V_2\t \C\pinv,
\label{eq:prediction-matrix-class-restrict}
\end{equation}
then the data driven scheme is conservative with respect to the optimal scheme. 

\end{enumerate}

\end{prop}\vspace{3pt}\noindent
See \Cref{sec:proof4} for a proof.

Before providing additional intuition as to the importance of this result, and the role of the filtering term $\F$, it is first demonstrated that the data driven scheme is a generalization of the truncated singular value decomposition calibration method.

\subsubsection{Truncated SVD Inversion}

The truncated singular value decomposition inversion or calibration method is a commonly used technique \cite{aster_parameter_2012, moore_role_2005, white_quantifying_2014} that generates the solution to the parameter estimation problem using a truncated decomposition of the data matrix. It will be demonstrated that the truncated SVD scheme can be replicated with the appropriate selection of a filtering matrix. To start, consider the SVD of the simplified data matrix 
\begin{equation}
\tG=\tU \tS \tV\t.
\end{equation}
Now, let $\tU_t$ denote the  columns of $\tU$ that correspond to the largest $k$ nonzero singular values. Furthermore, consider the filtering matrix defined by
\begin{equation} 
\F_{\text{TSVD}} \define \tU_t\t  .
\end{equation}
With this definition, the filtered data matrix $\tG'=\F_{\text{TSVD}} \tG$ becomes
\begin{equation}
\tG' = \F_{\text{TSVD}} \tG = \tU_t\t \tU \tS \tV\t = \tS_t \tV_t \t  ,
\end{equation}
where $\tV_t$, $\tS_t$ similarly denote truncated versions of $\tV$, $\tS$. Now, as $\F_{\text{TSVD}} \tG$ has full row rank (only nonzero singular values are included), the results of \cref{prop:data-driven-performance}(1) allow the estimator matrix for the scheme to be given by the pseudoinverse of $\tG' = \F_{\text{TSVD}} \tG$ and thus can be written as
\begin{equation}
\E\dat_{\text{TSVD}} = \tV_t \tS_t\i  .
\end{equation}
The estimated parameter vector of the simple model is now related to the data vector via the inverse of the truncated data matrix $\tG$, i.e.
\begin{equation}
\tmean{\v|\d'} \define \E\dat_{\text{TSVD}} \d' = \tV_t \tS_t\i \tU_t\t \d .
\end{equation}
This demonstrates that the truncated SVD inversion method can be replicated with the selection of an appropriate filtering matrix.

It is important to note however that this does not mean that the predictions generated by the truncated SVD method will be conservative. To guarantee this, the prediction must have the form specified in condition \cref{eq:prediction-matrix-class-restrict} of \cref{prop:data-driven-performance}, which in this case requires the prediction matrix to have the form
\begin{equation}
\Y = \A \tU_t \G + \B \tV_{\bar{t}} \t \C\pinv  ,
\end{equation}
where $\tV_{\bar{t}}$ contains the columns of $\tV$ that were removed by the truncation. If this is not obeyed, the predictions may be over confident. Furthermore, this condition is implicitly dependent on the truncation point $k$, such that this may hold for some values and not others. This dependency explains in part the results of the simulation studies performed by \citet{white_quantifying_2014} which demonstrated the difficulty in choosing an appropriate truncation point such that predictive uncertainty is accurately estimated by a simplified model.

\subsubsection{Selection of Data Filtering}

A key element of the defined data driven prediction scheme is the filtering matrix $\F$. For the results of \cref{prop:data-driven-performance} to hold, and the scheme to be conservative, two main requirements must be met:
\begin{enumerate}
\item $\F$ must be selected such that $\tG'  = \F\tG$ has linearly independent rows. This is a fairly trivial requirement, and simply requires that dependent, e.g. duplicated, measurements be combined, e.g. by averaging. 
Note that the information content is preserved as the error covariance is also transformed. 
In addition, it requires measurements that are insensitive to the model parameters be dropped.
\item $\F$ must be chosen such that the induced separation of the parameter vector $\v$ into those components that are estimatable $\v_1=\V_1 \v$ and lie in the rowspace of  $\F\tG$ from those that are not $\v_2=\V_2 \v$ and lie in the nullspace of $\F\tG$, forces all components of the parameter vector that have a direct influence on the prediction to be contained in the vector $\v_2$ and are thus not updated during the calibration process. This ensures that these parameters do not take on any surrogate roles, and that the predictions are not corrupted.
\end{enumerate}
In addition to these two requirements, the filtering matrix can be selected to improve the predictive performance by removing data components that are only weakly informative to the parameters. This can be considered identical to selecting an appropriate truncation point on a SVD calibration scheme such that the highly informative prior knowledge can be used instead of data that only imposes weak constraints, e.g. as recommended by \citet{moore_role_2005}. 

However before searching for an optimal filter, it is important to note that for a given type of data, predictions, and simplification, defined by the triple $\G, \Y, \C$, there may not be an $\F$ that allows the prediction to be cast in the form required by \cref{prop:data-driven-performance} and the use of the data driven scheme is not guaranteed to be conservative. This reinforces the fact that choosing a model simplification and performing calibration with a particular dataset, must be considered an integrated task that is explicitly dependent on the types of predictions that are required.

\subsubsection{Summary}

\begin{figure}[tb]%
\centering
\includegraphics[width=0.6\columnwidth]{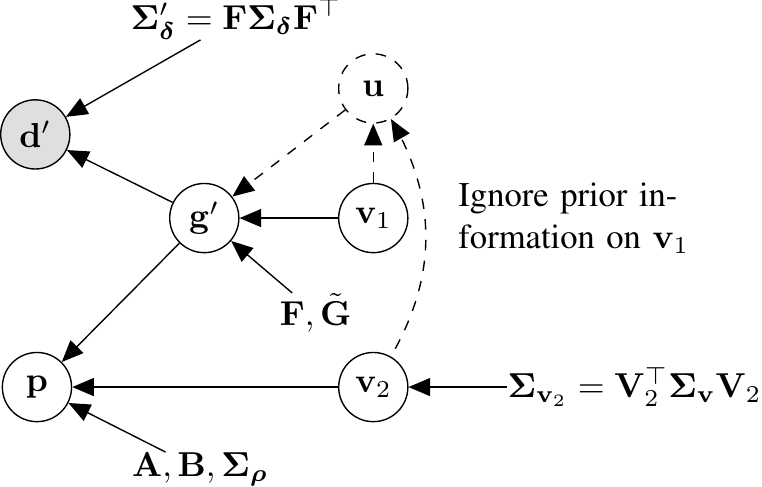}
\caption{Bayesian network depicting the structure of the densities $\ap\dat(\v)$, $\ap\dat(\d|\v)$, and $\ap\dat(\p|\v)$ employed by the data driven scheme. Also displayed in dashed lines are the dependencies on unmodelled complexity $\u$ of a suboptimal model under condition \cref{eq:prediction-matrix-class-restrict} of \cref{prop:data-driven-performance}. Note also that the simplified prediction matrix is given by $\tY=\A\tG' + \B\V_2\t$, however the actual values of $\A$ and $\B$ are not needed.}%
\label{fig:data-driven-scheme-BN}%
\end{figure}

This section has introduced a calibration and prediction scheme that is a generalization of the typically used truncated SVD scheme. Furthermore it has been demonstrated that the scheme is guaranteed to produce a conservative prediction posterior when the unmodelled complexity is hidden behind the data. The general structure of the prediction problem, that obeys this condition, is depicted in graphically in \cref{fig:data-driven-scheme-BN}. This represents the probability densities that define the data driven scheme and includes the dependency structure of the predictions (as required by condition \cref{eq:prediction-matrix-class-restrict} and depicted in \cref{fig:predict-class-BN-2}) as well as the structure of the inflated prior covariance matrix $\tcovdata{\v}$ (as specified in \cref{eq:data-driven-inflation}).

The important contribution here is the derived structural condition between the data, predictions and the simplifications that determines when it is adequate to use the data driven scheme, with a simplified model, to assess predictive uncertainty. It is noted that unlike the optimally compensated scheme, this scheme can be used without needing the explicit numerical values within the data and prediction matrices of the high fidelity reference model. It is noted however that, structural information from them is still required to ensure \cref{prop:data-driven-performance} is met. Although this still may not be straightforward in practice, structural considerations are generally significantly easier to handle than numerical ones.

\section{Discussion and Conclusions}
\label{sec:conclusion}

This paper has considered what constitutes a simplified, but useful, model. In particular it has examined how simplified models can be used to combine data and expert knowledge within a calibration or inversion process to generate a prediction with a conservative estimate of uncertainty. The concept of an optimal simplified model was defined that determines when standard probabilistic calibration methods are adequate to quantify predictive uncertainty. The main contribution is the introduction of two new calibration and prediction schemes, along with conditions that explicitly define when they are appropriate to generate a conservative estimate of uncertainty, for suboptimal models. These conditions explicitly relate the nature of the calibration data, the predictions of interest as well as the simplifications within the model. 

The first scheme allows the optimal posterior distribution to be generated by allowing the simplifications to be overcome by adjusting how the beliefs of the modeller are used to define a prior term. The scheme is only applicable for highly parameterized models. Furthermore, it requires the prior covariance for the parameters of the simple model to be generated using the data and prediction matrices of a high fidelity reference model. This is a significant limitation as if these matrices could be generated for a practical problem they could be directly used  to produce a prediction posterior without the need to consider a simplified model. Nevertheless, the value of this scheme lies in defining the ideal calibration process for a simplified model and demonstrates it is possible to overcome suboptimal simplifications through the judicial selection of a prior regularization matrix.

The second data driven scheme is designed for predictions that are strongly related to the data, such that the unmodelled complexity effects both in the same way. This scheme does not require the data and prediction matrices of the high fidelity model to be available for model calibration in the way the optimally compensated scheme does. However, it is only guaranteed to be conservative if the predictions, data and simplification have the required structural form.  The key insight provided by this scheme is understanding how the model simplifications can be overcome with the use of the right calibration data. 

It was also demonstrated that this data driven scheme is a  generalization of the popular truncated singular value decomposition inversion scheme \cite{aster_parameter_2012}. The generalization allows greater flexibility in filtering the data to ensure that the predictions are of the appropriate form for a given simplified model. Both calibration schemes have been applied to a prototypical groundwater prediction problem in \Cref{sec:example}.

Finally, it is noted that each of the two newly defined calibration and prediction schemes have conditions that are linked to the data and prediction matrices of a high fidelity reference model. 
For any practical problem it is unlikely that these conditions can be directly assessed  and further subjective judgment will be needed. To make this process easier, two areas of further work can be pursued. Firstly, more synthetic experimental analyses are required to demonstrate how the two schemes can be applied in more complex problems, e.g. as in \cite{white_quantifying_2014}. The second area is to understand the effect of partial non-satisfaction of these conditions, and determine how the schemes can be made robust to these.

It is also noted that the use of Bayesian networks, for instance as depicted in Figure \ref{fig:example-BN}, may be of great benefit to determine when the required structural conditions are likely to be obeyed, for instance, when is a given simplified model optimal, or when is it appropriate to use the data driven scheme. Such structural considerations may also help frame the arguments put forward in modelling projects (e.g. performed within environmental impact assessment studies) as to why a given modelling and calibration approach is adequate for a given prediction problem. For instance these argument can be framed using a two stage process. The first may conceptually link the complex system to an optimally simplified model, this stage would consider the features and characteristics of the system that ideally should be modelled. The second stage may then put forward arguments as to how any further simplifications used to produce a suboptimal numerical model will be handled by the calibration process.

Lastly, several other important areas of further research are identified:
\begin{itemize}

\item \textbf{Nonlinear simulators.} The analysis within this paper has required a linear relationship between the system properties and the data and predictions. This has allowed generic insights to be obtained, but is a significant limitation and further work is needed to relax this. One approach is to consider higher order expansions of the models such that some of the nonlinearities can be included, for example second order expansions have been considered in \cite{box_bias_1971, mclaughlin_distributed_1988, cooley_bias_2006}. Alternatively, more direct probabilistic formulations may also be possible that e.g. generalize the data driven scheme and only exploit the structural constraints with the problem. 

\item \textbf{Nonlinear parameterizations.} In the developed analytical framework only linear relationships between the parameters of the high fidelity model and the simplified model where considered. This should be extended to considered nonlinear parameterizations. %

\item \textbf{Over constrained calibration problems.} The results obtained for the optimally compensated prediction scheme required that the simplified model is under constrained by the data and provides no insight to aid the calibration of over constrained problems. Nevertheless, other approaches may be possible that reproduce the optimal, or at least a conservative, result. Similar modifications may also be possible for the data driven scheme.

\item \textbf{Conservativeness.} This paper has employed a definition of conservativeness based on the mean and covariance of the model predictions. Generalizations of this to non-Gaussian distributions have been considered in e.g. \cite{bailey_conservative_2012, ajgl_conservativeness_2013}. However, it is perhaps more appropriate to consider a generalization of conservativeness that explicitly includes the subsequent decision problem (i.e. engineering design or environmental management). This can be performed using a decision theoretic approach \cite{berger_statistical_1985} that includes the utility function of this subsequent decision problem and defines an approximate density $\ap(\p)$ as conservative if the expected utility is not over estimated, e.g.
\begin{equation*}
E_{\ap(\p)} \{ U(\mathbf{a},\p) \} \leq  E_{p(\p)} \{ U(\mathbf{a},\p) \} \quad \text{for all $\mathbf{a}$}.
\end{equation*}
Here $U(\mathbf{a},\p)$ is the utility function that encodes the gain under action $\mathbf{a}$, when the consequence  $\p$ occurs. Such a generalization would also allow the incorporation of the decision problem into how modelling and calibration should be performed. 
It is noted that the above generalization is equivalent to \cref{defn:conservative} when $U(\mathbf{a},\p)$ has the form of a weighted squared difference, i.e. $U(\mathbf{a},\p) = -(\mathbf{a}-\p)\t \A (\mathbf{a}-\p)$ for arbitrary positive semidefinite matrix $\A$.

\end{itemize}

\newpage
\appendix

\section{Example: Groundwater Head Prediction}
\label{sec:example}

\begin{figure}[tb]
\centering
\includegraphics[width=0.9\columnwidth]{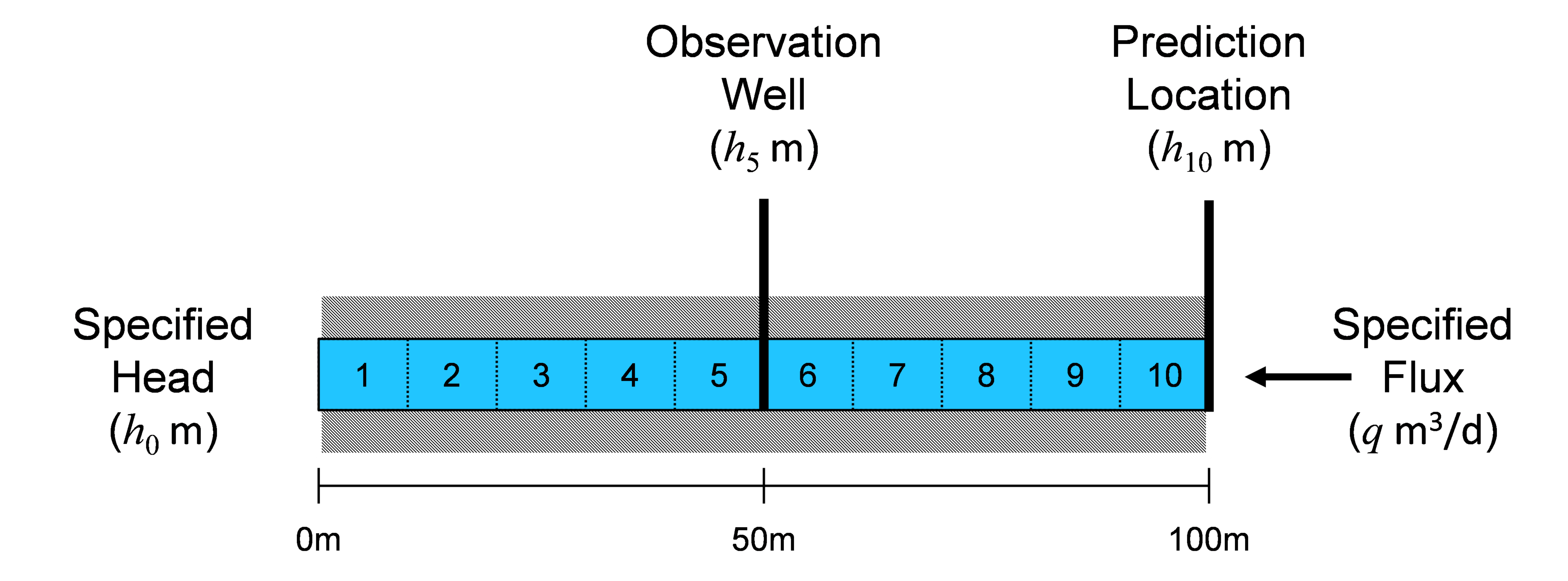}%
\caption{Groundwater prediction problem of interest. The aquifer is considered to be unit width into the page.}%
\label{fig:1d-aquifer}%
\end{figure}

The two newly defined calibration and prediction schemes are now applied to a prototypical groundwater prediction problem similar to that considered in \cite{white_quantifying_2014}. The scenario is depicted in \cref{fig:1d-aquifer} and consists of a 1D confined aquifer with a single observation well. Of interest is the predicted hydraulic head within the aquifer to the right of the observation well. It is noted that this is a very rudimentary problem, however  the objective here is to demonstrate the differences between the schemes and how the conditions that guarantee conservativeness relate to a specific problem.

\subsection{Prior Beliefs}

For this example, the expert modeller has the following beliefs about the system (taken from \cite{white_quantifying_2014} where possible)

\begin{enumerate}

\item A constant head boundary condition is believed to exist at the left side, corresponding to a discharge location. The head at this location, $h_0$, is believed to be normally distributed with mean $1.0$m and standard deviation $0.75$m above the upper confining layer.
\item No areal recharge or leakage is believed to take place within the domain of interest.
  
\item The thickness of the aquifer $b$ is known to be a constant over the domain with value $10$m.

\item The rate $q$ of water flowing through the aquifer known to be $0.5$m$^3$/day.

\item The system is believed to be in steady state. Furthermore, a numerical  simulation of the system with cell length $\ell = 10$m is believed to be adequate to capture the spatial variation in the head field that is of interest.

\item The hydraulic conductivity of the aquifer is believed to be heterogeneous, with a mean value of $2.5$m/day. A set of 10 cells, each of length $\ell = 10$m, is considered adequate to describe the heterogeneity, where the hydraulic conductivity of each cell, $K_i$, is believed to be log normally distributed such that $\log_{10} K_i$ has a mean $\log_{10} 2.5 \approx 0.398$, and a spatial correlation described by the exponential variogram with sill of $0.1\approx (0.316)^2$ and range $300$m.

\item The error in the data acquisition method is believed to be small and normally distributed with a mean of zero and standard deviation $0.1$m.
\end{enumerate}
It is noted that these capture what is \emph{believed} to realistically describe the system (or at least an optimally simplified model). Further simplifying \emph{assumptions} will be considered in \Cref{sec:example-assumed-simplifications}. 

Based on the above prior knowledge, the system properties can be defined as the vector that captures the constant boundary head, and the set of hydraulic conductivities
\begin{equation}
\x \define [h_0,\log_{10}(K_1),\dotsc,\log_{10}(K_{10})]\t
\end{equation}
Furthermore, the uncertainty in the system properties is fully captured by the mean and covariance matrix that defines the Gaussian distribution $p(\x) = N(\x;\mean{\x},\cov{\x})$.

To define the data and prediction likelihood functions, consider the vector $\x$ as known and define the mean with the following nonlinear functions, corresponding to the finite difference solution to Darcy's equations \cite{bear_modeling_2010} %
\begin{align}
\GG(\x) &= h_0 + \sum_{i=1}^5 \frac{q \ell}{b K_i},
\\
\YY(\x) &= h_0 + \sum_{i=1}^{10}  \frac{q \ell}{b K_i},
\\
&= \GG(\x) +  \sum_{i=6}^{10} \frac{q \ell}{b K_i}.
\label{eq:groundwater-prediction-dd}
\end{align}
Note the variables $\ell$, $b$ and $q$ are all known constants.
Furthermore, as the numerical simulation at this discretization is believed to be adequate, the data error $\nd$ is completely captured by errors in the measurement process such that $\cov{\nd}=(0.1)^2$. In addition the prediction error variance is taken to be zero, $\cov{\np} = 0$.

It is noted that the prediction equation has been rewritten on line \cref{eq:groundwater-prediction-dd} to explicitly include the data equation as an additive term. This form will be important when judging whether the data driven scheme is appropriate for a particular system simplification.

\subsection{Data Generation}

The measured head at the observation well is $\d=2.5$m. This is generated by the data equations $\d = \GG(\x_t)$ with no added measurement error. In addition the system properties $\x_t$ are the same as the prior mean $\mean{\x}$, with the exception that the boundary head is set to $h_0 = 1.5$m. This difference corresponds to $2/3$ of the prior standard deviation.

\subsection{Linearized Solution}

From the nonlinear functions $\GG(\x)$ and $\YY(\x)$, a pair of linear functions are constructed by linearizing about the prior mean with a first order Taylor expansion \cite{mclaughlin_reassessment_1996, carrera_inverse_2005}. 
\begin{align*}
\GG(\x) &\approx \GG(\mean{\x}) + \nabla_{\x}\GG( \mean{\x} ) [\x - \mean{\x}]
\end{align*}
and similarly for $\YY$. Furthermore, a transform into zero mean increments is performed such that  $\Dx = \x - \mean{\x}$, $\Dd = \d - \GG(\mean{\x})$, and $\Dp = \p - \YY(\mean{\x})$. Under the linearized approximation the data and prediction equations become
\begin{align*}
\Dd &\approx \G\Dx + \nd
\\
\Dp &\approx \Y\Dx + \np
\end{align*}
where the data and prediction matrices are defined by the Jacobian matrices evaluated at the prior mean $\G=\nabla_{\x}\GG( \mean{\x} )$ and  $\Y=\nabla_{\x}\YY( \mean{\x} )$.

The solutions to the full non-linear prediction problem and the approximate linearized version defined above are given in \cref{fig:example-prediction-schemes}(a). The non-Gaussian posterior density of the nonlinear problem has been produced using a standard Metropolis MCMC algorithm \cite{brooks_handbook_2011} to generate samples that are then smoothed using a kernel density estimator \cite{scott_multivariate_1992}. Overlaid with this is the Gaussian posterior produced from the linearized problem. It is noted that the posterior under the non-linear equations is slightly more peaked and non-symmetric when compared to the posterior generated by the linearized scheme.

It is noted that the non-linearity is only present as the log conductivity is considered to be normally distributed. If, for instance, the hydraulic resistance (inverse conductivity) was represented directly, the equations would become linear. However, this is not pursued here and the error caused by assuming a linear approximation of the nonlinear equations is considered out of scope.

\subsection{Assumed Simplifications}
\label{sec:example-assumed-simplifications}

Consider the following simplifying assumptions
\begin{enumerate}
\item[A1:] The constant flow boundary condition is assumed known and set to the prior value of $1.0$m. 
\item[A2:] Cells 1-5 are grouped together and assumed a homogeneous unit. This set of cells will be referred to as Zone A.
\item[A3:] Cells 6-10 are grouped together and assumed a homogeneous unit. This set of cells will be referred to as Zone B.
\end{enumerate}
This enables the parameters of the simple model to be defined by the vector of log conductivities of the two zones
\begin{equation*}
\v = [\log_{10} K_A, \log_{10} K_B]
\end{equation*}
This simplification scheme corresponds to a $\C$ matrix of 
\begin{equation*}
\C =\matrix{ 
0 & 0\\[-1pt]
1 & 0\\[-4pt]
\vdots & \vdots\\[-2pt]
1 & 0\\[-1pt]
0 & 1\\[-4pt]
\vdots & \vdots\\[-2pt]
0 & 1
}.
\end{equation*}
The use of increments allows the state of the high fidelity model to be defined in terms of an increment $\Dv$ such that
\begin{equation*}
\Dx = \C \Dv.
\end{equation*}
Furthermore, the simplified data equation becomes
 \begin{align*}
\tGG(\Dv) &= \GG(\mean{\x} + \C\Dv)
\end{align*}
and similarly for the prediction equation. Now, the linearized form of the simplified data and prediction equations become
\begin{align*}
\Dd &\approx \tG\Dv + \nd
\\
\Dp &\approx \tY\Dv + \np
\end{align*}
Here the simplified data and prediction matrices can be written in terms of the high fidelity model $\tG = \G\C$ and $\tY = \Y\C$, where $\G$ and $\Y$ are the Jacobian matrices of the high fidelity model $\G=\nabla_{\x}\GG( \mean{\x} )$,  $\Y=\nabla_{\x}\YY( \mean{\x} )$.

It is now of interest to determine whether this simplification is optimal. If it is, the naive prediction scheme when used with the simplified model will produce the same results as the optimal scheme applied to the high fidelity model. 
For the simplification to be optimal, the unmodelled complexity should have no effect on the data or predictions. This occurs when the following conditions hold
\begin{equation*}
\G\D = 0 \quad \text{and} \quad \Y\D = 0
\end{equation*}
where the columns of $\D$ define an orthonormal basis for the cokernel of $\C$, as defined in \cref{eq:param-subspace}. The unmodelled complexity consists of $h_0$ and the two 4D parameters that describe the small scale heterogeneity in the conductivity of the two zones. It is noted that this condition does not hold and thus the simplification is not optimal. Furthermore, the failure of this condition to hold is completely due to assumption A1 that considers the boundary condition $h_0$ is known. 
If this assumption is removed and $h_0$ included in the simplified model, while at the same time retaining the homogeneity assumptions of A2 and A3, then the simplification would become optimal (under the linearization considered). 
This optimal simplification will not be considered and it will be expected that the naive method will not be conservative.

\begin{figure}[tb]%
\centering
\begin{tabular}{c@{\hskip 2mm}c}
\includegraphics[width=0.41\columnwidth]{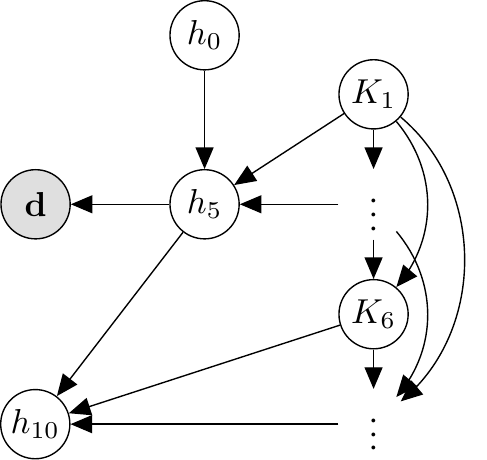}
&
\includegraphics[width=0.51\columnwidth]{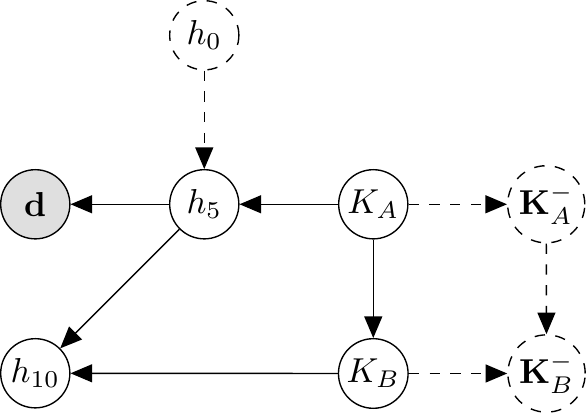}
\\
(a) & (b)
\end{tabular}
\caption{Bayesian Network that depicts the structure of the problem with (a) the reference model (b) the simplified model. The unmodelled complexity in the simplified model includes the boundary head $h_0$ and the small scale complexity within each zone, denoted by the two random vectors $\mathbf{K}_A^-$ and $\mathbf{K}_B^-$, each with four elements. The dependencies these have are denoted by dashed lines. }%
\label{fig:example-BN}%
\end{figure}

Finally, the structure of the high fidelity prediction problem and the simplified problem is depicted graphically with two Bayesian networks in \cref{fig:example-BN}. It is noted that the suboptimality of the simplification can be observed in \cref{fig:example-BN}(b) as the data and predictions are not conditionally independent of the unmodelled components ($h_0$) given the model parameters ($K_A$ and $K_B$).

\subsection{Prediction Results}

\begin{figure}[tb]%
\centering
\begin{tabular}{c@{}c}
\includegraphics[width=0.48\columnwidth]{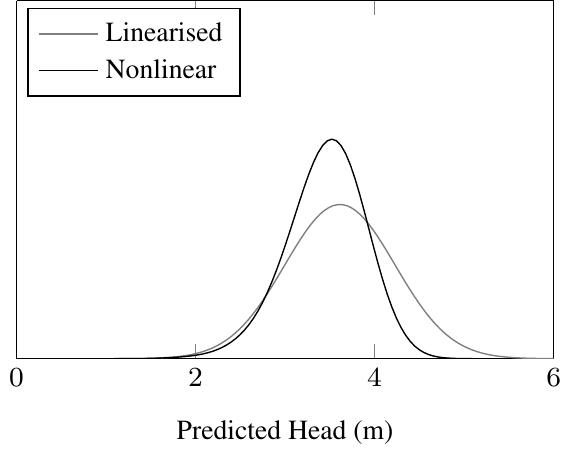}
&
\includegraphics[width=0.48\columnwidth]{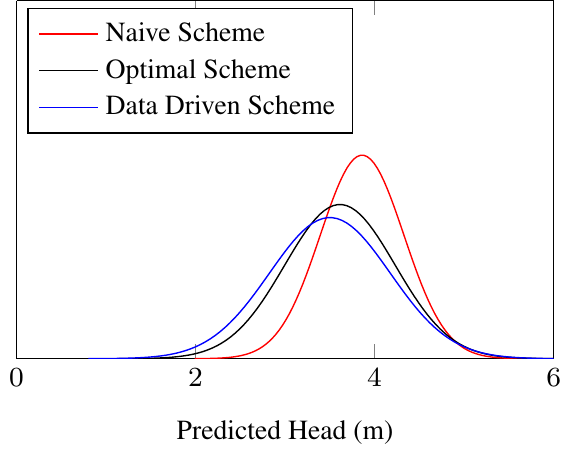}
\\
(a) & (b) 
\end{tabular}
\caption{Prediction probability density functions under (a) reference model and (b) the simplified model. The optimally compensated scheme in (b) reproduces the posterior distribution of the linearized version of the reference model in (a). The naive scheme is non-conservative and under estimates the uncertainty, while the data driven scheme is conservative.}%
\label{fig:example-prediction-schemes}%
\end{figure}

Here, the performance of the two new calibration and prediction schemes are considered, with the resulting posterior density functions for the predicted head displayed in \cref{fig:example-prediction-schemes}(b). The results are also compared with the naive scheme which represents a typical probabilistic calibration scheme of a simplified model. It is important to note that no ``ground truth'' value for the prediction is given for comparison purposes as the objective is to produce the full probability distribution and not make a single point prediction.

\subsubsection{Naive Scheme}

The naive scheme considers the uncertainty in the parameters included by the simple model, but ignores any errors introduced in the simplification. As the simplification is suboptimal, it should not be expected that the prediction posterior will be conservative. Furthermore, the results in \cref{fig:example-prediction-schemes}(b) demonstrate that this is true for this scenario as the posterior is overly narrow.

To understand the cause of the overconfidence in this scenario, consider the prior covariance in the parameters $\Dv$ as defined in \cref{prop:uncertainty-propagation} 
\begin{equation}
\cov{\v} = \C\pinv \cov{\x} \C\pinv {}\t 
\approx 10^{-2}\matrix{
    8.58&    6.19 \\
    6.19  &  8.58} 
\end{equation}
Due to the large range of 300m used for the variogram of the log conductivity field, the correlation between the two parameters maintained by the simple model is considerable. In addition, the known flow rate and the assumption of known boundary head $h_0$, allows the head measurement to be used to very accurately, but incorrectly, estimate the conductivity $K_A$. Furthermore, due to the large correlation that $K_A$ has with $K_B$, this scheme produces an overconfident prediction probability density function for the parameter $K_B$, which in turn causes the predicted head to be also overconfident and non-conservative.

\subsubsection{Optimally Compensated Scheme}

Now consider the optimally compensated calibration and prediction scheme. Firstly it is noted that the simplified problem is highly parameterized as $D_\v \geq D_\p + D_\d$ as $D_\v=2$ and $D_\p=D_\d=1$. The major difference between the naive scheme and the optimally compensated scheme is the modification of the prior covariance used for the parameters of the simple model. In particular, the modified prior of this scheme is not just  defined in terms of the simplification matrix, but also in terms of the data and prediction matrices, via the matrix $\R = \tZ\pinv \Z$, where $\Z=\tmatrix{\G \\ \Y}$ and $\tZ=\tmatrix{\tG \\ \tY}$. The optimally compensating prior covariance becomes
\begin{equation*}
\tcovopta{\v} =  \R\cov{\x}\R\t
\approx 10^{-2}\matrix{
    27.44  &  6.19 \\
    6.19  &  8.58}
\end{equation*}
It is noted that the only difference between the values of the naive prior covariance $\cov{\v}$ and $\tcovopta{\v}$, is the inflation of the marginal variance of the first component corresponding to $\log K_A$. This causes less information to be propagated from the data into $\log K_B$, and enables the prediction posterior to replicate the results of the high fidelity model.

The key difficulty with this scheme is that it requires the data and prediction matrices of the high fidelity model in order to produce the inflated prior covariance matrix $\tcovopta{\v}$ that compensates for the suboptimal simplifications.

\subsubsection{Data Driven Scheme}

To apply the data driven calibration and scheme, it is first of interest to determine if it is appropriate for the problem, in particular, does the prediction have the right structural relationship with the data and the simplifications as specified in \cref{prop:data-driven-performance}. This will be checked first using the algebraic expression, and then using the structural characteristics of the network depicted in \cref{fig:example-BN}(b). 

The required algebraic condition requires that for some $\A$ and $\B$ the prediction matrix can be written as
\begin{equation}
\Y = \A\F\G + \B \V_2\t\C\pinv.
\label{eq:gw-dd-condition}
\end{equation}
For this problem, the data is scalar, and thus the filtering matrix will be set to unity $\F=1$. 
\newcommand{\T}{\mathbf{T}}
Also, from the prediction equation in \cref{eq:groundwater-prediction-dd}, it is clear that the prediction matrix $\Y$ is equal to the data matrix $\G$ plus an additional term, which will be denoted by $\T$. This allows the Jacobian matrix $\Y$ to be written as
\begin{equation}
\Y = \G + \T.
\end{equation}
Thus, the matrix $\A$ in \cref{eq:gw-dd-condition} will be set to unity $\A=1$.

Now it is of interest to consider the second term in \cref{eq:gw-dd-condition}. For the prediction to have the required form, the matrix $\T=\Y-\G$ must be equivalent to $\B \V_2\t\C\pinv$, and thus must satisfy two conditions:
\begin{gather}
(\Y-\G)\D = 0
\label{gw-dd-condition-1}
\\
(\Y-\G)\C\V_1 = (\tY - \tG)\V_1 = 0
\label{gw-dd-condition-2}
\end{gather}
The first condition \eqref{gw-dd-condition-1} requires the difference between the data and the prediction to be independent of the unmodelled complexity. For this scenario this can be checked algebraically using the high fidelity prediction and data matrices and is satisfied. Now consider the second condition \eqref{gw-dd-condition-2}, this requires the difference between the data and the prediction to be independent on the rowspace components of the $\tG$. This can be easily checked using just the simplified data and prediction matrices and is also satisfied.
Thus, it is guaranteed that the data driven scheme is conservative for this problem.  

As an alternative to this algebraic check, the conditional independence requirements can be check using the structure of the Bayesian network. In particular, the requirement of \cref{eq:gw-dd-condition} is equivalent to the requirement that the prediction $\p=h_{10}$ is conditionally independent of all unmodelled components, $\u=[h_0, \mathbf{K}_A^-, \mathbf{K}_B^-]$, given the uncorrupted data $\g = h_5$ and the nullspace components of the parameter vector $\v_2=K_B$. From the structure of the network, these two nodes are the only parents of the prediction and thus, the  required independence requirement is satisfied. This graphical approach provides much greater intuition into when the requirements are met.

When the data driven scheme is applied, it is noted that the available prior knowledge is modified in two ways. Firstly, the prior information on $K_A$ is ignored and this parameter is estimated from the data only. Secondly, the correlation between $K_A$ and $K_B$ is also ignored, this causes the posterior over the parameter $K_B$ to be the same as the prior. These modifications are encapsulated in the data driven prior covariance matrix
\begin{equation*}
\tcovdata{\v} =  
\lim_{\alpha \rightarrow \infty}
\alpha \V_1 \V_1\t +  \V_2\V_2\t\cov{\v}\V_2\V_2\t
=
10^{-2}\matrix{
    \infty  &  0 \\
    0  &  8.58}.
\end{equation*}
The posterior generated by this scheme is conservative and is depicted in \cref{fig:example-prediction-schemes}(b). It is noted that the posterior is slightly shifted than that produced by the optimal scheme and has a larger variance.

\subsection{Summary}

This section has demonstrated the operation of the naive, optimally compensated, and data driven calibration and prediction schemes on a prototypical groundwater problem. A high fidelity model was considered to represent the true belief of an modeller and a suboptimal simplification developed to represent a computational model. 

The key difference between the three schemes is in the specification of the prior covariance matrix for the parameters of the simplified model. In particular, the naive scheme generates this by directly projecting the prior distribution that exists in the parameters of the high fidelity model into the parameters of the simplified model. The optimally compensated scheme performs the projection using full knowledge of the data and prediction matrices of the high fidelity model. Lastly, the data driven scheme uses the same matrix as the naive scheme but throws away some of the information it contains.

The conditions under which the schemes are conservative (or optimal) have also been highlighted. The naive scheme should only be applied to optimally simplified models, this was not satisfied and the posterior shown to be overconfident and non-conservative. The optimally compensated scheme should only be applied to highly parameterized models. And the data driven scheme is conservative only when the predictions and data have a similar dependency to the unmodelled complexity. These last two conditions were shown to hold for the problem considered, and the generated posterior distributions were also conservative.
\section[Proof of Proposition]{Proof of Proposition 1}
\label{sec:proof1}

\begin{proof}
For a given simplification matrix, the parameters of the high fidelity model $\x$ can be expanded using \cref{eq:state-decomposition} as
\begin{equation*}
\x = \C \v + \D \u = \matrix{\C & \D} \matrix{\v \\ \u}.
\end{equation*}
Furthermore, from the decomposition in \cref{eq:param-subspace}, the matrix $\matrix{\C & \D}$ can be rewritten as
\begin{equation*}
\matrix{\C & \D} = \matrix{\Ucol{\C} & \D} \matrix{\Sing{\C} \Vrow{\C}\t & 0 \\0 &  \I}.
\end{equation*}
This is nonsingular and has an inverse that simplifies to
\begin{align*}
\matrix{\C & \D}\i = \matrix{ \C\pinv \\  \D\t},
\end{align*}
where $\C\pinv = \Vrow{\C} \Sing{\C}\i \Ucol{\C}\t$. Thus, the transformed covariance matrix in the space of $\v$, $\u$ has the form
\begin{equation*}
\matrix{\cov{\v} & \cov{\v\u} \\ \cov{\v\u}\t & \cov{\u}} = \matrix{ \C\pinv \\  \D\t} \cov{\x} \matrix{ \C\pinv{}\t &  \D}.  
\end{equation*}
\end{proof}

\section[Proof of Proposition]{Proof of Proposition 2}
\label{sec:proof2}

\begin{proof}
Under the conditions of part (1) the matrix $\C$ denotes an optimal simplification and thus $\G\D=\Y\D=0$, where $\D$ is an orthonormal basis for the cokernel of $\C$, e.g. as given in \cref{eq:param-subspace}. 
Using the expansion of the parameter vector as $\x=\C\v + \D\u$, the covariance $\cov{\x}$ becomes
\begin{equation}
\cov{\x} = \C\cov{\v}\C\t  + \D\cov{\u}\D\t + \C\cov{\v\u}\D\t  + \D\cov{\v\u}\t\C\t.
\label{eq:cov-x-expansion}
\end{equation}
Substituting \cref{eq:cov-x-expansion} into the posterior mean and covariance of the optimal scheme, defined in \cref{eq:pred-update} and \cref{eq:predcov-update}, and noting that $\G\D=\Y\D=0$,  the following forms are obtained
\begin{align*}
\mean{\p|\d} 
&= \Y \C\cov{\v} \C\t\G\t (\G \C\cov{\v} \C\t \G\t + \cov{\nd})^{-1} \d,  
\\
\cov{\p|\d} 
&= \Y \C\cov{\v} \C\t \Y\t + \cov{\np} 
\\ & \quad
- \Y \C\cov{\v} \C\t \G\t (\G \C\cov{\v} \C\t \G\t + \cov{\nd})^{-1} \G \C\cov{\v} \C\t \Y\t.   
\end{align*}
Noting that $\tG=\G\C$ and $\tY=\Y\C$ these expressions are equivalent to mean and covariance those of the naive scheme, i.e. $\tmeannaivea{\p|\d} = \mean{\p|\d}$  and $\tcovnaivea{\p|\d} = \cov{\p|\d}$ for all $\d$. This proves part (1) of the proposition.

To proceed, to parts (2) and (3), for the naive scheme to be conservative, the mean and covariance of the posterior must obey \cref{eq:conservative-scheme}, which simplifies to
\begin{equation}
\tcov{\p|\d}
\succeq 
E_{p(\d,\p)} \left\{  
 (\tmeannaivea{\p|\d}-\p)(\tmeannaivea{\p|\d}-\p)\t 
\right\}
\label{eq:conservative1111}
\end{equation}
Also, the expectation over $\d,\p$ is equivalent to an expectation over the independent variables $\x,\nd,\np$, where $\d = \G\x+\nd$ and $\p = \Y\x+\np$. Thus to prove part (2) of the proposition, it will be now shown that this holds under the special condition of $\Y\D = \tY \E\naive\G\D$, defined in \cref{eq:naive-balanced-error}. 

To start, note that $\tmeannaivea{\p|\d} = \tY\E\naive\d$, $\d=[\G\x+\nd]$ and $\x=\C\v + \D\u$, thus the difference $\tmeannaivea{\p|\d}-\p$ can be written  
\begin{align}
\tmeannaivea{\p|\d}-\p & = [\tY\E\naive\tG -\tY]\v + \tY\E\naive\nd - \np + [\tY\E\naive\G\D -\Y\D]\u
\label{eq:mean-diff-full2222}
\\
& = [\tY\E\naive\tG -\tY]\v + \tY\E\naive\nd - \np.
\label{eq:mean-diff-opt-simp}
\end{align}
Where, \cref{eq:mean-diff-opt-simp} has used condition \cref{eq:naive-balanced-error}.
Now, as $\v=\C\pinv\x,\nd,\np$ are all independent and zero mean, the expected squared difference of $\tmeannaivea{\p|\d}-\p$ can be written in terms of the covariance matrices
\begin{align*}
& E \left\{  
 (\tmeannaivea{\p|\d}-\p)(\tmeannaivea{\p|\d}-\p)\t 
\right\} 
\\
&\quad\quad\quad = 
[\tY\E\naive\tG -\tY] \cov{\v} [\tY\E\naive\tG -\tY]\t 
+ \tY \E\naive \cov{\nd} \E\naive{}\t \tY\t + \cov{\np}.
\end{align*}
Expanding the right hand side with the naive estimator matrix $\E\naive \define \cov{\v} \tG\t (\tG \cov{\v} \tG\t + \cov{\nd})^{-1}$ and simplifying, produces the result
\begin{align*}
& E \left\{  
 (\tmeannaivea{\p|\d}-\p)(\tmeannaivea{\p|\d}-\p)\t 
\right\} 
\\
&\quad\quad\quad = 
\tY\cov{\v} \tY\t + \cov{\np} 
- \tY \cov{\v} \tG\t (\tG \cov{\v} \tG\t + \cov{\nd})^{-1} \tG \cov{\v} \tY\t 
\\
&\quad\quad\quad = \tcovnaivea{\p|\d}
\end{align*}
Thus, the special condition $\Y\D = \tY \E\naive\G\D$ is sufficient for  \cref{eq:conservative1111} to be satisfied (with equality), ensuring the naive scheme is conservative. This proves part (2).

To prove part (3), consider equation \cref{eq:mean-diff-full2222} above and assume $\v$ and $\u$ are independent and $\tY\E\naive\G\D \neq \Y\D$, then the expected squared difference of $\tmeannaivea{\p|\d}-\p$ becomes
\begin{multline*}
E \left\{  
 (\tmeannaivea{\p|\d}-\p)(\tmeannaivea{\p|\d}-\p)\t 
\right\}  = 
\\
[\tY\E\naive\tG -\tY] \cov{\v} [\tY\E\naive\tG -\tY]\t 
+ \tY \E\naive \cov{\nd} \E\naive{}\t \tY\t + \cov{\np}
\\
+ [\tY\E\naive\G\D -\Y\D] \cov{\u} [\tY\E\naive\G\D -\Y\D]\t .
\end{multline*}
Defining $\M=\tY\E\naive\G\D -\Y\D \neq 0$, this simplifies to 
\begin{equation*}
E \left\{  
 (\tmeannaivea{\p|\d}-\p)(\tmeannaivea{\p|\d}-\p)\t 
\right\}  = 
\tcovnaivea{\p|\d} 
+  \M \cov{\u} \M\t.
\end{equation*}
Now, as $\M$ is a non-zero matrix and $\cov{\u}$ is a non-zero positive semi-definite covariance matrix, $\M\cov{\u}\M\t$ must be non-zero and positive semi-definite. Thus, the required condition \cref{eq:conservative1111} does not hold as
\begin{equation*}
E \left\{  
 (\tmeannaivea{\p|\d}-\p)(\tmeannaivea{\p|\d}-\p)\t 
\right\}  \succ 
\tcovnaivea{\p|\d}.
\end{equation*}
and the scheme is strictly not conservative. This proves part (3).
\end{proof}

\section[Proof of Proposition]{Proof of Proposition 3}
\label{sec:proof3}

\begin{proof}
To prove optimality of the compensated scheme it is necessary to show the conditions defined by \cref{eq:optimal-inference-conditions} are satisfied by the prior covariance matrix $\tcovopta{\v}$ defined by this scheme. 
It is noted that these are satisfied when 
\begin{equation*}
\tG\tcovopta{\v}\tG\t = \G\cov{\x}\G\t, \quad
\tY\tcovopta{\v}\tY\t = \Y\cov{\x}\Y\t, \quad \text{and} \quad
\tY\tcovopta{\v}\tG\t = \Y\cov{\x}\G\t.
\end{equation*}
These conditions can be combined and rewritten as
\begin{equation}
\tZ\tcovopta{\v}\tZ\t  = \Z\cov{\x}\Z\t ,
\label{eq:55555}
\end{equation}
where $\Z = \tmatrix{\G\\ \Y}$ and $\tZ = \tmatrix{\tG\\ \tY}$. Now with the definition of $\tcovopta{\v} =  \tZ\pinv \Z\cov{\x}\tZ\pinv{}\t \Z\t$, condition \cref{eq:55555} becomes
\begin{equation*}
\tZ\tZ\pinv \Z \cov{\x} \tZ\pinv{}\t \Z\t\tZ\t  = \Z\cov{\x}\Z\t ,
\end{equation*}
and holds when $\tZ\tZ\pinv \Z = \Z$.

Now, to demonstrate $\tZ\tZ\pinv \Z = \Z$ is satisfied, the condition $\rank(\Z\C) = \rank(\Z)$
 implies that $\range(\tZ) = \range(\Z\C) = \range(\Z)$ \cite[3.16]{seber_matrix_2008}. 
Furthermore, it is noted that $\range(\tZ)=\range(\tZ\pinv{}\t)$ \cite[7.52(l)]{seber_matrix_2008}, and thus
\begin{equation*}
\range(\Z) = \range(\tZ\pinv{}\t).
\end{equation*}
This implies that $\Z\t(\I - \tZ \tZ\pinv) = 0$ \cite[2.34]{seber_matrix_2008}. 
Furthermore, as $\tZ$ is real, $(\tZ \tZ\pinv)\t = \tZ \tZ\pinv$. Thus, taking the transpose of $\Z\t(\I - \tZ \tZ\pinv) = 0$, produces
\begin{equation*}
(\I - \tZ \tZ\pinv)\Z = 0
\end{equation*}
and hence $\tZ \tZ\pinv\Z = \Z$. 
\end{proof}

\section[Proof of Proposition]{Proof of Proposition 4}
\label{sec:proof4}

\begin{proof}
To start, note that the inverse of the prior covariance matrix $\tcovdata{\v}$ simplifies as follows
\begin{align}
[\tcovdata{\v}]\i &= \lim_{\alpha \rightarrow \infty} [\V_2\V_2\t\cov{\v}\V_2\V_2\t + \alpha \V_1 \V_1\t]\i
\\
&  = \V_2(\V_2\t\cov{\v}\V_2)\i\V_2\t
\label{eq:inv-expansion-111}
\end{align}
Now, the estimator matrix is defined as $\E\dat = \tcovdata{\v} \tG'{}\t (\tG' \tcovdata{\v} \tG'{}\t + \cov{\nd}')^{-1}$. Using the matrix inversion identity \cite[15.1(a)]{seber_matrix_2008}, the inverse definition in \cref{eq:inv-expansion-111}, the compact SVD of $\tG' = \U_1 \S_1  \V_1\t$, and noting that as $\tG'$ has full row rank $\U_1 \i = \U_1\t$, the estimator matrix simplifies to
\begin{align*}
\E\dat 
&=[ [\tcovdata{\v}]\i + \tG'{}\t \cov{\nd}'{}\i \tG'  ]\i \tG'{}\t \cov{\nd}'{}\i 
\\
&= \V_1 \S_1\i \U_1\t = \tG' {}\pinv.
\end{align*}
This can be used to directly define the mean of the prediction posterior 
\begin{align*}
\tmeandata{\p|\d} &= \predMean{}(\tcovdata{\v},\tG',\cov{\nd}',\tY,\cov{\np},\d')  
\\
&= \tY\E\dat \d'.
\end{align*}
With similar substitutions as above, and using the Woodbury identity \cite[15.3(b)(i)]{seber_matrix_2008}, the covariance of the prediction posterior simplifies to
\begin{align*}
\tcovdata{\p|\d} &= \predCov{}(\tcovdata{\v},\tG',\cov{\nd}',\tY,\cov{\np}) 
\\
&= \tY \tcovdata{\v} \tY\t + \cov{\np} - \tY \tcovdata{\v} \tG'{}\t (\tG\tcovdata{\v} \tG'{}\t + \cov{\nd}')^{-1} \tG' \tcovdata{\v} \tY\t   
\\
&= \tY [ [\tcovdata{\v}]\i + \tG'{}\t \cov{\nd}'{}\i \tG'{}  ]\i \tY\t    + \cov{\np} 
\\
&= \tY [ \V_2\V_2\t\cov{\v}\V_2\V_2\t + \V_1 \S_1\i \U_1\t \cov{\nd}' \U_1 \S_1\i \V_1\t  ] \tY\t 
 + \cov{\np} 
\\
&= \tY \W\t\cov{\v}\W \tY\t + \tY \E\dat \cov{\nd}' \E\dat{}\t \tY\t    + \cov{\np} 
\end{align*}
where $\W=\V_2\V_2\t=\I-\V_1\V_1\t$ represents a projection on to the nullspace of $\tG'$. This concludes the proof of part (1) of the proposition.

For the scheme to be conservative, the mean and covariance of the posterior must obey
\begin{equation}
\tcovdata{\p|\d}
\succeq 
E_{p(\d,\p)} \left\{  
 (\p-\tmeandata{\p|\d})(\p-\tmeandata{\p|\d})\t 
\right\}
\label{eq:conservative2222}
\end{equation}
Also, the expectation over $\d,\p$ is equivalent to an expectation over the independent variables $\x,\nd,\np$, where $\d = \G\x+\nd$ and $\p = \Y\x+\np$.

Now, the difference between the prediction $\p \define \Y\x + \np $ and the mean $\tmeandata{\p|\d}$ simplifies to
\begin{align*}
\p - \tmeandata{\p|\d}  &= \Y\x + \np - \tY\E\dat\F \d,
\\
&= [\tY  - \tY \E\dat \tG']\v  - \tY \E\dat \nd  + \np
+ [\Y\D- \tY \E\dat\G'\D]\u
\\
&= \tY \W \v - \tY \E\dat \nd  + \np
\end{align*}
Furthermore, as $\v=\C\pinv\x$, $\nd$ and $\np$ are all independent and zero mean, then the difference $\p - \tmeandata{\p|\d}$ has an expected value of zero and a covariance given which is equivalent to $\tcovdata{\p|\d}$
\begin{align*}
E \left\{(\p - \tmeandata{\p|\d})(\p - \tmeandata{\p|\d})\t \right\} &
\\
&\hspace{-2cm} = \tY \W \cov{\v} \W \tY\t + \tY \E\dat \cov{\nd}\E\dat{}\t \tY\t  + \cov{\np}
\\
&\hspace{-2cm}=\tcovdata{\p|\d}.
\end{align*}
Thus, the required condition \cref{eq:conservative2222} is satisfied (with equality) and the prediction scheme is conservative. This proves part (2) of the proposition.
\end{proof}

\bibliographystyle{siamplain_nourl}

\bibliography{SIAM_Water}

\begin{thebibliography}{10}

\bibitem{ajgl_conservativeness_2013}
{\sc J.~Ajgl and M.~Simandl}, {\em On conservativeness of posterior density
  fusion}, in 2013 16th {International} {Conference} on {Information} {Fusion}
  ({FUSION}), July 2013, pp.~85--92.

\bibitem{aster_parameter_2012}
{\sc R.~C. Aster, B.~Borchers, and C.~H. Thurber}, {\em Parameter {Estimation}
  and {Inverse} {Problems}, {Second} {Edition}}, Academic Press, Waltham, MA, 2
  edition~ed., Feb. 2012.

\bibitem{bailey_conservative_2012}
{\sc T.~Bailey, S.~Julier, and G.~Agamennoni}, {\em On conservative fusion of
  information with unknown non-{Gaussian} dependence}, in 2012 15th
  {International} {Conference} on {Information} {Fusion} ({FUSION}), July 2012,
  pp.~1876--1883.

\bibitem{bear_modeling_2010}
{\sc J.~Bear and A.~H.-D. Cheng}, {\em Modeling groundwater flow and
  contaminant transport}, Springer, Dordrecht; London, 2010.

\bibitem{berger_statistical_1985}
{\sc J.~Berger}, {\em Statistical {Decision} {Theory} and {Bayesian}
  {Analysis}}, Springer, New York, 2nd edition~ed., 1985.

\bibitem{beven_concept_2005}
{\sc K.~Beven}, {\em On the concept of model structural error}, Water Science
  and Technology: A Journal of the International Association on Water Pollution
  Research, 52 (2005), pp.~167--175.

\bibitem{beven_equifinality_2001}
{\sc K.~Beven and J.~Freer}, {\em Equifinality, data assimilation, and
  uncertainty estimation in mechanistic modelling of complex environmental
  systems using the {GLUE} methodology}, Journal of Hydrology, 249 (2001),
  pp.~11--29, \url{https://doi.org/10.1016/S0022-1694(01)00421-8}.

\bibitem{box_bias_1971}
{\sc M.~J. Box}, {\em Bias in {Nonlinear} {Estimation}}, Journal of the Royal
  Statistical Society. Series B (Methodological), 33 (1971), pp.~171--201.

\bibitem{brooks_handbook_2011}
{\sc S.~Brooks}, {\em Handbook for {Markov} chain {Monte} {Carlo}}, Taylor \&
  Francis, Boca Raton, 2011.

\bibitem{carrera_inverse_2005}
{\sc J.~Carrera, A.~Alcolea, A.~Medina, J.~Hidalgo, and L.~J. Slooten}, {\em
  Inverse problem in hydrogeology}, Hydrogeology Journal, 13 (2005),
  pp.~206--222, \url{https://doi.org/10.1007/s10040-004-0404-7}.

\bibitem{clark_unraveling_2006}
{\sc M.~P. Clark and J.~A. Vrugt}, {\em Unraveling uncertainties in hydrologic
  model calibration: {Addressing} the problem of compensatory parameters},
  Geophysical Research Letters, 33 (2006), p.~L06406,
  \url{https://doi.org/10.1029/2005GL025604}.

\bibitem{cooley_bias_2006}
{\sc R.~L. Cooley and S.~Christensen}, {\em Bias and uncertainty in
  regression-calibrated models of groundwater flow in heterogeneous media},
  Advances in Water Resources, 29 (2006), pp.~639--656,
  \url{https://doi.org/10.1016/j.advwatres.2005.07.012}.

\bibitem{craig_bayesian_2001}
{\sc P.~S. Craig, M.~Goldstein, J.~C. Rougier, and A.~H. Seheult}, {\em
  Bayesian {Forecasting} for {Complex} {Systems} {Using} {Computer}
  {Simulators}}, Journal of the American Statistical Association, 96 (2001),
  pp.~717--729, \url{https://doi.org/10.1198/016214501753168370}.

\bibitem{doherty_use_2011}
{\sc J.~Doherty and S.~Christensen}, {\em Use of paired simple and complex
  models to reduce predictive bias and quantify uncertainty}, Water Resources
  Research, 47 (2011), pp.~n/a--n/a,
  \url{https://doi.org/10.1029/2011WR010763}.

\bibitem{doherty_groundwater_2013}
{\sc J.~Doherty and C.~T. Simmons}, {\em Groundwater modelling in decision
  support: reflections on a unified conceptual framework}, Hydrogeology
  Journal, 21 (2013), pp.~1531--1537,
  \url{https://doi.org/10.1007/s10040-013-1027-7}.

\bibitem{doherty_short_2010}
{\sc J.~Doherty and D.~Welter}, {\em A short exploration of structural noise},
  Water Resources Research, 46 (2010),
  \url{https://doi.org/10.1029/2009WR008377}.

\bibitem{ferdowsian_explaining_2001}
{\sc R.~Ferdowsian, D.~J. Pannell, C.~McCarron, A.~Ryder, and L.~Crossing},
  {\em Explaining groundwater hydrographs: separating atypical rainfall events
  from time trends}, Soil Research, 39 (2001), pp.~861--876.

\bibitem{freeze_hydrogeological_1990}
{\sc R.~A. Freeze, J.~Massmann, L.~Smith, T.~Sperling, and B.~James}, {\em
  Hydrogeological {Decision} {Analysis}: 1. {A} {Framework}}, Ground Water, 28
  (1990), pp.~738--766,
  \url{https://doi.org/10.1111/j.1745-6584.1990.tb01989.x}.

\bibitem{goldstein_probabilistic_2004}
{\sc M.~Goldstein and J.~Rougier}, {\em Probabilistic {Formulations} for
  {Transferring} {Inferences} from {Mathematical} {Models} to {Physical}
  {Systems}}, SIAM Journal on Scientific Computing, 26 (2004), pp.~467--487,
  \url{https://doi.org/10.1137/S106482750342670X}.

\bibitem{goldstein_bayes_2006}
{\sc M.~Goldstein and J.~Rougier}, {\em Bayes {Linear} {Calibrated}
  {Prediction} for {Complex} {Systems}}, Journal of the American Statistical
  Association, 101 (2006), pp.~1132--1143,
  \url{https://doi.org/10.1198/016214506000000203}.

\bibitem{goodwin_quantifying_1992}
{\sc G.~C. Goodwin, M.~Gevers, and B.~Ninness}, {\em Quantifying the error in
  estimated transfer functions with application to model order selection}, IEEE
  Transactions on Automatic Control, 37 (1992), pp.~913--928,
  \url{https://doi.org/10.1109/9.148344}.

\bibitem{goodwin_stochastic_1989}
{\sc G.~C. Goodwin and M.~E. Salgado}, {\em A stochastic embedding approach for
  quantifying uncertainty in the estimation of restricted complexity models},
  International Journal of Adaptive Control and Signal Processing, 3 (1989),
  pp.~333--356, \url{https://doi.org/10.1002/acs.4480030405}.

\bibitem{gupta_towards_2012}
{\sc H.~V. Gupta, M.~P. Clark, J.~A. Vrugt, G.~Abramowitz, and M.~Ye}, {\em
  Towards a comprehensive assessment of model structural adequacy}, Water
  Resources Research, 48 (2012), p.~W08301,
  \url{https://doi.org/10.1029/2011WR011044}.

\bibitem{hunt_are_2007}
{\sc R.~J. Hunt, J.~Doherty, and M.~J. Tonkin}, {\em Are {Models} {Too}
  {Simple}? {Arguments} for {Increased} {Parameterization}}, Ground Water, 45
  (2007), pp.~254--262, \url{https://doi.org/10.1111/j.1745-6584.2007.00316.x}.

\bibitem{kennedy_bayesian_2001}
{\sc M.~C. Kennedy and A.~O'Hagan}, {\em Bayesian calibration of computer
  models}, Journal of the Royal Statistical Society: Series B (Statistical
  Methodology), 63 (2001), pp.~425--464,
  \url{https://doi.org/10.1111/1467-9868.00294}.

\bibitem{ljung_model_1999}
{\sc L.~Ljung}, {\em Model {Validation} and {Model} {Error} {Modeling}}, in
  Proceedings of the {Asrom} {Symposium} on {Control}, Lund, Sweden, 1999,
  pp.~15--42.

\bibitem{ljung_system_1999}
{\sc L.~Ljung}, {\em System {Identification}: {Theory} for the {User}},
  Prentice Hall, Upper Saddle River, NJ, 2 edition~ed., Jan. 1999.

\bibitem{ljung_perspectives_2010}
{\sc L.~Ljung}, {\em Perspectives on system identification}, Annual Reviews in
  Control, 34 (2010), pp.~1--12,
  \url{https://doi.org/10.1016/j.arcontrol.2009.12.001}.

\bibitem{ljung_stochastic_2014}
{\sc L.~Ljung, G.~C. Goodwin, and J.~C. Agüero}, {\em Stochastic {Embedding}
  {Revisited}: {A} {Modern} {Interpretation}}, nation, 15 (2014), p.~35.

\bibitem{mclaughlin_reassessment_1996}
{\sc D.~McLaughlin and L.~R. Townley}, {\em A {Reassessment} of the
  {Groundwater} {Inverse} {Problem}}, Water Resources Research, 32 (1996),
  pp.~1131--1161, \url{https://doi.org/10.1029/96WR00160}.

\bibitem{mclaughlin_distributed_1988}
{\sc D.~McLaughlin and E.~F. Wood}, {\em A distributed parameter approach for
  evaluating the accuracy of groundwater model predictions: 2. {Application} to
  groundwater flow}, Water Resources Research, 24 (1988), pp.~1048--1060,
  \url{https://doi.org/10.1029/WR024i007p01048}.

\bibitem{moore_role_2005}
{\sc C.~Moore and J.~Doherty}, {\em Role of the calibration process in reducing
  model predictive error}, Water Resources Research, 41 (2005), p.~W05020,
  \url{https://doi.org/10.1029/2004WR003501}.

\bibitem{ninness_estimation_1995}
{\sc B.~Ninness and G.~C. Goodwin}, {\em Estimation of model quality},
  Automatica, 31 (1995), pp.~1771--1797,
  \url{https://doi.org/10.1016/0005-1098(95)00108-7}.

\bibitem{poeter_all_2007}
{\sc E.~Poeter}, {\em All {Models} are {Wrong}, {How} {Do} {We} {Know} {Which}
  are {Useful}?}, Ground Water, 45 (2007), pp.~390--391,
  \url{https://doi.org/10.1111/j.1745-6584.2007.00350.x}.

\bibitem{poeter_mma_2007}
{\sc E.~P. Poeter and M.~C. Hill}, {\em {MMA}, {A} computer code for
  {Multi}-{Model} {Analysis}}, U.{S}. {Geological} {Survey} {Techniques} and
  {Methods} 6-E3, 2007.

\bibitem{reichert_analyzing_2009}
{\sc P.~Reichert and J.~Mieleitner}, {\em Analyzing input and structural
  uncertainty of nonlinear dynamic models with stochastic, time-dependent
  parameters}, Water Resources Research, 45 (2009), p.~W10402,
  \url{https://doi.org/10.1029/2009WR007814}.

\bibitem{reinelt_comparing_2002}
{\sc W.~Reinelt, A.~Garulli, and L.~Ljung}, {\em Comparing different approaches
  to model error modeling in robust identification}, Automatica, 38 (2002),
  pp.~787--803, \url{https://doi.org/10.1016/S0005-1098(01)00269-2}.

\bibitem{rojas_application_2010}
{\sc R.~Rojas, S.~Kahunde, L.~Peeters, O.~Batelaan, L.~Feyen, and
  A.~Dassargues}, {\em Application of a multimodel approach to account for
  conceptual model and scenario uncertainties in groundwater modelling},
  Journal of Hydrology, 394 (2010), pp.~416--435,
  \url{https://doi.org/10.1016/j.jhydrol.2010.09.016}.

\bibitem{rougier_probabilistic_2007}
{\sc J.~Rougier}, {\em Probabilistic {Inference} for {Future} {Climate} {Using}
  an {Ensemble} of {Climate} {Model} {Evaluations}}, Climatic Change, 81
  (2007), pp.~247--264, \url{https://doi.org/10.1007/s10584-006-9156-9}.

\bibitem{rougier_uncertainty_2014}
{\sc J.~Rougier and M.~Crucifix}, {\em Uncertainty in climate science and
  climate policy}, arXiv:1411.6878 [physics],  (2014).
\newblock arXiv: 1411.6878.

\bibitem{scott_multivariate_1992}
{\sc D.~W. Scott}, {\em Multivariate {Density} {Estimation}: {Theory},
  {Practice}, and {Visualization}}, Wiley, New York, 1st edition~ed., Aug.
  1992.

\bibitem{seber_matrix_2008}
{\sc G.~A.~F. Seber}, {\em A matrix handbook for statisticians},
  Wiley-Interscience, Hoboken, N.J., 2008.

\bibitem{smith_bayesian_2010}
{\sc J.~Q. Smith}, {\em Bayesian {Decision} {Analysis}: {Principles} and
  {Practice}}, Cambridge University Press, Sept. 2010.

\bibitem{strong_when_2014}
{\sc M.~Strong and J.~Oakley}, {\em When {Is} a {Model} {Good} {Enough}?
  {Deriving} the {Expected} {Value} of {Model} {Improvement} via {Specifying}
  {Internal} {Model} {Discrepancies}}, SIAM/ASA Journal on Uncertainty
  Quantification, 2 (2014), pp.~106--125,
  \url{https://doi.org/10.1137/120889563}.

\bibitem{tarantola_inverse_2005}
{\sc A.~Tarantola}, {\em Inverse {Problem} {Theory} and {Methods} for {Model}
  {Parameter} {Estimation}}, Society for Industrial and Applied Mathematics,
  Philadelphia, PA, 1 edition~ed., 2005.

\bibitem{von_asmuth_modeling_2008}
{\sc J.~R. Von~Asmuth, K.~Maas, M.~Bakker, and J.~Petersen}, {\em Modeling
  {Time} {Series} of {Ground} {Water} {Head} {Fluctuations} {Subjected} to
  {Multiple} {Stresses}}, Ground Water, 46 (2008), pp.~30--40,
  \url{https://doi.org/10.1111/j.1745-6584.2007.00382.x}.

\bibitem{voss_editors_2011-1}
{\sc C.~I. Voss}, {\em Editors message: {Groundwater} modeling fantasies - part
  1, adrift in the details}, Hydrogeology Journal, 19 (2011), pp.~1281--1284,
  \url{https://doi.org/10.1007/s10040-011-0789-z}.

\bibitem{voss_editors_2011}
{\sc C.~I. Voss}, {\em Editors message: {Groundwater} modeling fantasies - part
  2, down to earth}, Hydrogeology Journal, 19 (2011), pp.~1455--1458,
  \url{https://doi.org/10.1007/s10040-011-0790-6}.

\bibitem{vrugt_improved_2005}
{\sc J.~A. Vrugt, C.~G.~H. Diks, H.~V. Gupta, W.~Bouten, and J.~M. Verstraten},
  {\em Improved treatment of uncertainty in hydrologic modeling: {Combining}
  the strengths of global optimization and data assimilation}, Water Resources
  Research, 41 (2005), p.~W01017, \url{https://doi.org/10.1029/2004WR003059}.

\bibitem{watson_parameter_2013}
{\sc T.~A. Watson, J.~E. Doherty, and S.~Christensen}, {\em Parameter and
  predictive outcomes of model simplification}, Water Resources Research,
  (2013), pp.~3952--3977, \url{https://doi.org/10.1002/wrcr.20145}.

\bibitem{white_quantifying_2014}
{\sc J.~T. White, J.~E. Doherty, and J.~D. Hughes}, {\em Quantifying the
  predictive consequences of model error with linear subspace analysis}, Water
  Resources Research, 50 (2014), pp.~1152--1173,
  \url{https://doi.org/10.1002/2013WR014767}.

\end{thebibliography}
\end{document}